\documentclass{llncs}
\usepackage{graphicx}
\usepackage{amsmath}
\usepackage{amssymb}
\usepackage[usenames,dvipsnames]{xcolor}
\usepackage{mathtools}
\usepackage{mathrsfs}
\usepackage{adjustbox}
\usepackage{booktabs}
\usepackage{tikz}
\usepackage{amstext} 
\usepackage{array}   
\usepackage[moderate,paragraphs=normal]{savetrees}

\usepackage[shortlabels]{enumitem} 
\usepackage[hidelinks]{hyperref}

\usepackage{wrapfig}

\usepackage{orcidlink}

\newcommand{\fc}[1]{{\color{teal}[FC:#1]}}
\newcommand{\np}[1]{{\color{blue}[NP:#1]}}

\newcommand{\HA}{\mathcal{M}}

\newcommand{\SM}{S(\mathcal{M})}
\newcommand{\FutureHorizon}{H_f}
\newcommand{\PastHorizon}{H_p}
\newcommand{\cdom}{\mathit{Var}}
\newcommand{\ddom}{\mathit{Loc}}
\newcommand{\cvar}{v}
\newcommand{\dvar}{l}
\newcommand{\state}{s}
\newcommand{\Flow}{Flow}
\newcommand{\Jump}{Jump}
\newcommand{\Trans}{Trans}
\newcommand{\RReset}{Reset}
\newcommand{\Inv}{Inv}

\newcommand{\ncf}{\delta}

\newcommand{\StateDistrib}{\mathcal{S}}

\newcommand{\mcal}{\mathcal}
\newcommand{\mbb}{\mathbb}

\newcommand{\credibility}{\kappa}
\newcommand{\sat}{\mathsf{Sat}}
\newcommand{\barsat}{\mathsf{SAT}}
\newcommand{\stochsat}{\mathsf{SSat}}

\newcommand{\OutS}{\mathsf{B}}

\newtheorem{myproblem}{Problem}

\begin{document}
\title{Learning-Based Approaches to Predictive Monitoring with Conformal Statistical Guarantees}

\author{Francesca Cairoli\inst{1}\orcidlink{0000-0002-6994-6553} \and
Luca Bortolussi \inst{1}\orcidlink{0000-0001-8874-4001}\and
Nicola Paoletti\inst{2}\orcidlink{0000-0002-4723-5363}
}

\institute{University of Trieste, Italy \and King's College London, United Kingdom}

\maketitle    

\vspace{-0.5cm}

\begin{abstract}
This tutorial focuses on efficient methods to predictive monitoring (PM), the problem of detecting at runtime future violations of a given requirement from the current state of a system. While performing model checking at runtime would offer a precise solution to the PM problem, it is generally computationally expensive. To address this scalability issue, several lightweight approaches based on machine learning have recently been proposed. These approaches work by learning an approximate yet efficient surrogate (deep learning) model of the expensive model checker.  A key challenge remains to ensure reliable predictions, especially in safety-critical applications.

We review our recent work on predictive monitoring, one of the first to propose learning-based approximations for CPS verification of temporal logic specifications and the first in this context to apply conformal prediction (CP) for rigorous uncertainty quantification. These CP-based uncertainty estimators offer statistical guarantees regarding the generalization error of the learning model, and they can be used to determine unreliable predictions that should be rejected.
In this tutorial, we present a general and comprehensive framework summarizing our approach to the predictive monitoring of CPSs, examining in detail several variants determined by three main dimensions: system dynamics (deterministic, non-deterministic, stochastic), state observability, and semantics of requirements' satisfaction (Boolean or quantitative).

\end{abstract}

\vspace{-0.75cm}

\section{Introduction}

\vspace{-0.2cm}

Verification of temporal properties for a cyber-physical systems (CPS) is of
paramount importance, especially with CPSs having become ubiquitous in safety-critical domains, from autonomous vehicles to medical devices~\cite{alur2015principles}.
We focus on \emph{predictive monitoring} (PM), that is, the problem of predicting, at runtime, if a safety violation is imminent from the current CPS state. In this context, PM has the advantage, compared to traditional monitoring~\cite{bartocci2018specification}, of detecting potential safety violations before they occur, in this way enabling preemptive countermeasures to steer the system back to safety (e.g. switching to a failsafe mode as done in the Simplex architecture~\cite{johnson2016real}). Thus, effective PM must balance between prediction accuracy, to avoid errors that can jeopardize safety, and computational efficiency, to support fast execution at runtime. 


We focus on correctness specifications given in Signal Temporal Logic (STL)~\cite{maler2004monitoring,donze2010robust}, a popular language for formal reasoning about CPS. An advantage of STL is that it admits two semantics, the usual Boolean semantics and a quantitative (robust) semantics, which quantifies the degree of satisfaction of a property. When using the latter, we speak of \textit{quantitative PM (QPM)}.


Performing model checking of STL requirements at run-time would provide a precise solution to the PM problem (precise up to the accuracy of the system's model), but such a solution is computationally expensive in general, and thus infeasible for real-world applications.
For this reason, a number of approximate PM techniques based on machine learning have been recently proposed (see e.g.~\cite{bortolussi2021neural,cairoli2021neural,cairoli2023conformal,lindemann2023conformal,muthali2023multi}).

In this paper, we review our work on learning-based methods for PM, developed under the name of \emph{Neural Predictive Monitoring}~\cite{bortolussi2019neural}.
The core idea is that when a satisfaction (SAT) oracle is at our disposal, we can approximate it using deep learning models trained using a set of oracle-labeled examples. 
The resulting learning-based model overcomes the scalability issues faced by the original oracle: a forward pass of a (reasonably sized) neural network is most often more efficient than computing satisfaction of an STL property, especially if the underlying system is non-deterministic or stochastic. 
However, such a solution is inherently approximate, and so it becomes essential -- especially in safety-critical domains -- to offer assurances regarding the generalization performance of our approximation. 
To this purpose, we rely on \textit{conformal prediction (CP)~\cite{vovk2005algorithmic}}, a technique that allows us to complement model predictions with uncertainty estimates that enjoy (finite-sample, i.e., non-asymptotic) statistical guarantees on the model's generalization error. 
CP requires only very mild assumptions on the data\footnote{The only assumption is exchangeability, a weaker version of the independent and identically distributed assumption. A collection of $N$ values is exchangeable if the $N!$ different orderings
are equally likely, i.e. have the same joint probability.} and it is flexible enough to be applied on top of most predictors. Furthermore, computing CP-based uncertainty estimates is highly efficient, meaning that our approach can offer statistical guarantees on the PM predictions without affecting performance and runtime applicability.


In this tutorial, we present a general and comprehensive framework summarizing several variants of the neural predictive monitoring approach, variants determined by the following three  dimensions: 
\begin{enumerate}
    \item \emph{Dynamics}: The CPS dynamics can be either deterministic, non-deterministic, or stochastic, 
    depending on whether the future behavior of the system is uniquely determined by its current state, is uncertain or exhibits randomness.
    \item \emph{State Observability}: The current CPS state can be fully observable, 
    or we may only have access to partial and noisy measurements of the state, making it more challenging to obtain accurate predictions of the CPS evolution. 
    \item \emph{Satisfaction}: The type of property satisfaction can be either Boolean or quantitative. In the former case, the PM outcome is a ``yes" or ``no" answer (the CPS either satisfies the property or does not). In the latter case, the outcome is a quantitative degree of satisfaction, which quantifies the robustness of STL (Boolean) satisfaction to perturbations (in space or time) of the CPS trajectories.   
\end{enumerate}
By considering the above dimensions, we can design accurate and reliable PM solutions in a variety of scenarios accounting for a vast majority of CPS models.


\vspace{-0.5cm}

\paragraph{Overview of the paper.} This paper is structured as follows. We start by presenting the background theory in Sect.~\ref{sec:preliminaries}. Sect.~\ref{sec:pm} states rigorous formalizations of the predictive monitoring problems. Sect.~\ref{sec:unc} provides
background on methods used to estimate predictive uncertainty and obtain statistical guarantees. Sect.~\ref{sec:metods} defines the different declinations of learning-based PM approaches. Related works are discussed in Sect.~\ref{sec:related}. Conclusions are drawn in Sect.~\ref{sec:conclusion}.

\vspace{-0.3cm}

\section{Background}\label{sec:preliminaries}

A cyber-physical system (CPS) is a system combining physical and digital components.
Hybrid systems (HS), whose dynamics exhibit both continuous and discrete dynamics, can well capture the mixed continuous and discrete behaviour of a CPS. 
An HS has both flows, described by differential equations, and jumps, described by a state machine or an automaton. The continuous behaviour depends on the discrete state and discrete jumps are similarly determined by the continuous state. 
The state of an HS is defined by the values of the continuous variables and by a discrete mode. 
The continuous flow is permitted as long as a so-called invariant holds, while discrete transitions can occur as soon as given jump conditions are satisfied. 

\begin{remark}
The approaches proposed in this paper are applicable to any black-box system for which a satisfaction (SAT) oracle is available. That said, HS represent a very useful and expressive class of models for which several SAT oracles (model checkers) have been developed. Therefore, we will focus on this class of models for the rest of the paper.
\end{remark}

A hybrid automaton (HA) is a formal model that mathematically describes the evolution in time of an HS. 
\begin{definition}[Hybrid automaton]\label{def:ha}
A \emph{hybrid automaton} (HA) is a tuple $\mathcal{M} = (\mathit{Loc,}$ $\mathit{Var,}$ $\mathit{Init, Flow, Trans, Inv})$, where 
$\mathit{Loc}$ is a finite set of discrete \emph{locations} (or \emph{modes}); $\mathit{Var} = \{ \cvar_1, \ldots, \cvar_n\}$ is a set of 
continuous \emph{variables}, evaluated over a \emph{continuous domain} $\cdom \subseteq \mathbb{R}^n$; $\mathit{Init} \subseteq S(\mathcal{M})$ is the set of \emph{initial states}, where $S(\mathcal{M}) = \mathit{Loc} \times \cdom$ is the \emph{state space} of $\mathcal{M}$; $\mathit{Flow: Loc} \to (\cdom \to \cdom)$ is the \emph{flow} function, defining the continuous dynamics at each location; $\mathit{Trans}$ is the \emph{transition relation}, consisting of tuples of the form $(l, g, r, l')$, where $l,l' \in \mathit{Loc}$ are \emph{source and target locations}, respectively, $g \subseteq \cdom$ is the \emph{guard}, and $r: \cdom \to \cdom$ is the \emph{reset}; $\mathit{Inv : Loc}\to 2^{\cdom}$ is the \emph{invariant} at each location.
\end{definition}
In general, HA dynamics can be either deterministic, non-deterministic, or stochastic. \emph{Deterministic} HAs are a special case where the transition relation is a function of the source location. On the other hand, when the continuous flow or the discrete transitions happen according to a certain probability distribution, we have a \emph{stochastic} HA. In this case, the dynamics is represented as the combination of continuous stochastic flows probability 
$\Flow :(\ddom\times\cdom)\to (\cdom\to [0,1])$ and discrete jump probability 
$\Trans: (\ddom\times 2^{\cdom}) \to ((\ddom\times(\cdom\times \cdom))\to[0,1])$. In particular, 
$\Flow(\cvar'\mid \dvar,\cvar)$ denotes the probability of a change rate of $\cvar'$ when in state $(\dvar,\cvar)$ and 
$\Trans(r,l'\mid l, g)$ is the probability of applying reset $r$ and jumping into $\dvar'$ through a transition starting from $\dvar$ with guard $g$. We often prefer to avoid transitions with both non-determinism and stochasticity and so, for each stochastic HA location, we define its invariant and guards so that they form a partition of $\cdom$.

We define a \emph{signal} as a function $\Vec{\state}:\mbb{T}\to \mbb{V}$, where $\mbb{T}\subset\mbb{R}^+$ is the time domain, whereas $\mbb{V}$ determines the nature of the signal. 
If $\mbb{V} = \mbb{B} :=\{true, false\}$, we have a \emph{Boolean signal}. If $\mbb{V} = \mbb{R}$, we have a \emph{real-valued signal}. 
We consider signals that are solutions of a given HA. 
Let $\tau = \{
[t_i,t_{i+1}] | t_i\le t_{i+1},i=1,2,\dots
\}$ 
be a hybrid time trajectory.
If $\tau$ is infinite the last interval may be open on the right. Let $\mbb{T}$ denote the set of hybrid time trajectories. Then, for a given $\tau\in\mbb{T}$, a \emph{hybrid signal} $\Vec{\state}:\tau\to \SM$ defined on $\tau$ with values in a generic hybrid space $\SM$ is a sequence of functions
$$
\Vec{\state} = \{\Vec{\state}_i:[t_i,t_{i+1}]\to \SM \mid [t_i,t_{i+1}]\in\tau\}.
$$
In practice, it can be seen as a pair of hybrid signals $\Vec{\cvar}:\tau\to\cdom$ and $\Vec{\dvar}:\tau\to\ddom$ with $\tau\in\mbb{T}$, such that $(\Vec{\dvar}_1(t_1),\Vec{\cvar}_1(t_1))\in \mathit{Init}$, and for any $[t_i,t_{i+1}]\in\tau$, $\Vec{\dvar}_i$ is constant and there exist $g$ and $r$ such that $(l_i,g,r,l_{i+1})\in \mathit{Trans}$, $\vec{v}_i(t_{i+1}^-) \in g$, and $\vec{v}_{i+1}(t_{i+1})=r(\vec{v}_i(t_{i+1}^-))$.
     Moreover, for every $ t\in[t_i,t_{i+1}]$, it must hold that $\Vec{\cvar}_i(t) \in \mathit{Inv}(\dvar_i)$ and
 $$
    \Vec{\cvar}_i(t) = \Vec{\cvar}_i(t_i)+\int_{t_i}^t\Flow\left(\Vec{\dvar}_i(t'),\Vec{\cvar}_i(t')\right) dt'.$$


When the HA is stochastic, hybrid signals must have a non-zero probability, that is, for every piece $i$ of the signal, there exists $g,r$ such that 
$\Trans(r,l_{i+1}\mid l_i, g)>0$, $\vec{v}_i(t_{i+1}^-) \in g$, and $\vec{v}_{i+1}(t_{i+1})=r(\vec{v}_i(t_{i+1}^-))$. Moreover, for every $t \in [t_i,t_{i+1})$, it must hold that $\mathit{Flow}(\dot{\vec{v}}_i(t) \mid l_i, \vec{v}_i(t))>0$, where $\dot{\vec{v}}_i(t)=\lim_{dt\to 0} (\vec{v}_i(t+dt)-\vec{v}_i(t))/\vec{v}_i(t)$.


\begin{remark}\label{remark:markov}
We note that HA induces Markovian dynamics, that is, the evolution of the HA depends only on the current state. We do not see this as a restriction as most systems of interest are Markovian or can be made so by augmenting the state space.
\end{remark}


\vspace{-0.5cm}

\subsection{Signal Temporal Logic (STL)}\label{subsec:stl}

Signal temporal logic (STL)~\cite{maler2004monitoring} was originally developed in
order to specify and monitor the behaviour of physical
systems, including temporal constraints between events. STL
allows the specification of properties of dense-time, real-valued signals, and the automatic generation of monitors for
testing these properties on individual simulation traces. 
The rationale of STL is to transform hybrid signals into Boolean ones, using predicates built on the following \emph{STL syntax}:
\begin{equation}\label{eq:stl_syntax}
    \varphi := true\ |\ \mu_{\ell}\ |\ \mu_g \ |\ \neg\varphi\ |\  \varphi\land\varphi\ |\ \varphi\ \mcal{U}_{[a,b]}\varphi,
\end{equation}
where $[a,b]\subseteq \mbb{T}$ is a bounded temporal interval. 
For a hybrid signal $\Vec{\state}[t]$, 
$\mu_g$ denotes atomic predicates over continuous variables, with $g:\mathit{Var}\to \mathbb{R}$, whereas $\mu_{\ell}$ denotes atomic predicates over discrete variables, with $\ell \in \mathit{Loc}$.
From this essential syntax, it is easy to define other operators, used to abbreviate the syntax in a STL formula: $false:=\neg true$, $\varphi \lor \psi := \neg(\neg\varphi\land\neg\psi)$, $\Diamond_{[a,b]} \varphi:= true\ \mcal{U}_{[a,b]}\varphi$ and $\square_{[a,b]} \varphi:= \neg\Diamond_{[a,b]}\neg\varphi$.


\paragraph{Boolean semantics.}
The satisfaction of a formula $\varphi$ by a signal $\Vec{\state}$ at time $t$ is defined as:
\begin{itemize}
   \item[-] $(\Vec{\state},t) \models \mu_g \iff g(\Vec{\cvar}[t])>0$; 
\item[-] $(\Vec{\state},t) \models \mu_{\ell} \iff \Vec{\dvar}[t] = \ell$;
   \item[-] $(\Vec{\state},t) \models \varphi_1 \land\varphi_2 \iff (\Vec{\state},t) \models\varphi_1\land(\Vec{\state},t) \models\varphi_2$;
   \item[-] $(\Vec{\state},t) \models\neg\varphi \iff \neg((\Vec{\state},t) \models\varphi))$;
   \item[-] $(\Vec{\state},t) \models\varphi_1 \mcal{U}_{[a,b]}\varphi_2 \iff \exists t'\in [t+a,t+b] \mbox{ s.t. }\\ (\Vec{\state},t') \models\varphi_2\land \forall t''\in [t,t'), (\Vec{\state},t'') \models\varphi_1$.

    \item[-] Eventually: $(\Vec{\state},t) \models \Diamond_{[a,b]}\varphi \iff \exists t'\in[t+a,t+b]  \mbox{ s.t. } (\Vec{\state},t') \models\varphi$;
    \item[-] Always: $(\Vec{\state},t) \models \square_{[a,b]}\varphi \iff \forall t'\in[t+a,t+b]  \quad (\Vec{\state},t') \models\varphi$.
\end{itemize}
Given formula $\varphi$ and a signal $\Vec{\state}$ over a bounded time interval, we can define the Boolean satisfaction signal as $\chi^\varphi(\Vec{\state},t)= 1$ if $(\Vec{\state}, t)\models\varphi$ and $\chi^\varphi(\Vec{\state},t)= 0$ otherwise.
Monitoring the satisfaction of a formula is done recursively, by computing $\chi^{\varphi_i}(\Vec{\state},\cdot)$ for each sub-formula $\varphi_i$ of $\varphi$. The recursion is performed by leveraging the tree structure of the STL formula, where each node represents a sub-formula, in an incremental fashion, so that the leaves are the atomic
propositions and the root represents the whole formula. Thus the procedure goes bottom-up from atomic predicated to the top formula.

\paragraph{Quantitative semantics.} 
The main kind of quantitative STL semantics is \textit{space robustness}, which quantifies how much a signal can be perturbed with additive noise before changing the truth value of a given property $\varphi$~\cite{donze2010robust}. It is defined as a function $R_\varphi$ such that:

\begin{itemize}
    \item[-] $R_{\mu_g} (\Vec{\state},t) = g(\Vec{\cvar}[t])$; 
    \item[-] $R_{\neg\varphi}(\Vec{\state},t) = -R_{\varphi}(\Vec{\state},t)$;
    \item[-] $R_{\varphi_1\land\varphi_2}(\Vec{\state},t) = \min (R_{\varphi_1}(\Vec{\state},t),R_{\varphi_2}(\Vec{\state},t))$;
    \item[-] $R_{\varphi_1 \mcal{U}_{[a,b]}\varphi_2}(\Vec{\state},t) = \underset{t'\in [t+a,t+b]}{\sup}\left(\min\left(R_{\varphi_2}(\Vec{\state},t'), \underset{t''\in [t,t']}{\inf}R_{\varphi_1}(\Vec{\state},t'')\right)\right) $.
\end{itemize}
The sign of $R_\varphi$ indicates the satisfaction status:
$R_\varphi(\Vec{\state},t)>0\Rightarrow (\Vec{\state},t)\models\varphi$ and $R_\varphi(\Vec{\state},t)<0\Rightarrow (\Vec{\state},t)\not\models\varphi$. The definition of $R_{\mu_{\ell}}$, i.e., the robustness of discrete atoms, is arbitrary as long as it returns a non-negative value when $\mu_{\ell}$ is true  and non-positive when $\mu_{\ell}$ is false (a common choice is returning $+\infty$ and $-\infty$, respectively).
As for the Boolean semantics, it is possible to automatically generate monitors for the quantitative semantics as well.
The algorithm follows a similar bottom-up approach over the syntax tree of the formula. 

Similarly, \emph{time robustness} capture the effect on the satisfaction of shifting events in time. The left and right time robustness of an STL formula $\varphi$ with respect to a trace $\Vec{\state}$ at time $t$ are defined inductively by letting:
\begin{align*}
Q^-_\varphi(\Vec{\state},t) &= \chi^\varphi (\Vec{\state},t)\cdot\max\{d\ge 0\ s.t.\ \forall t'\in [t-dt],\ \chi^\varphi (\Vec{\state},t') = \chi^\varphi (\Vec{\state},t)\}\\
Q^+_\varphi(\Vec{\state},t) &= \chi^\varphi (\Vec{\state},t)\cdot\max\{d\ge 0\ s.t.\ \forall t'\in [t,t+d],\ \chi^\varphi (\Vec{\state},t') = \chi^\varphi (\Vec{\state},t)\}.
\end{align*}

While space and robustness are most common, our PM approach can support any other kind of STL quantitative semantics, e.g., based on a combined space-time robustness~\cite{donze2010robust} or resiliency~\cite{chen2022stl}. Hereafter, we represent a generic STL monitor, encompassing either Boolean or quantitative (spatial or temporal) satisfaction, as $C_\varphi\in\{\chi_\varphi, R_\varphi, Q_\varphi\}$.

\begin{figure}[!t]
    \centering
    \hspace{-5cm}
    \includegraphics[scale=0.095]{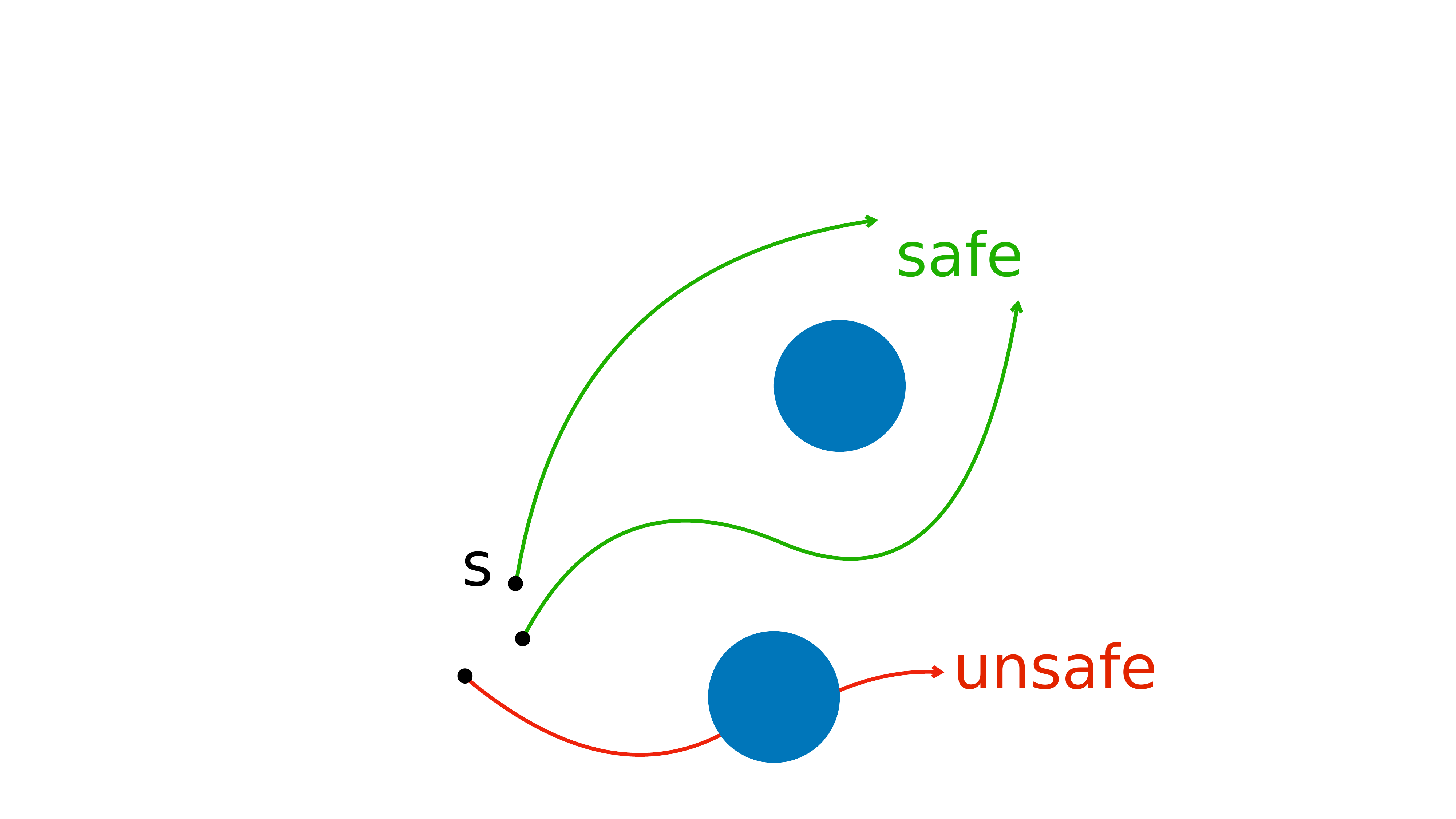}
    \hspace{-1cm}
    \includegraphics[scale=0.095]{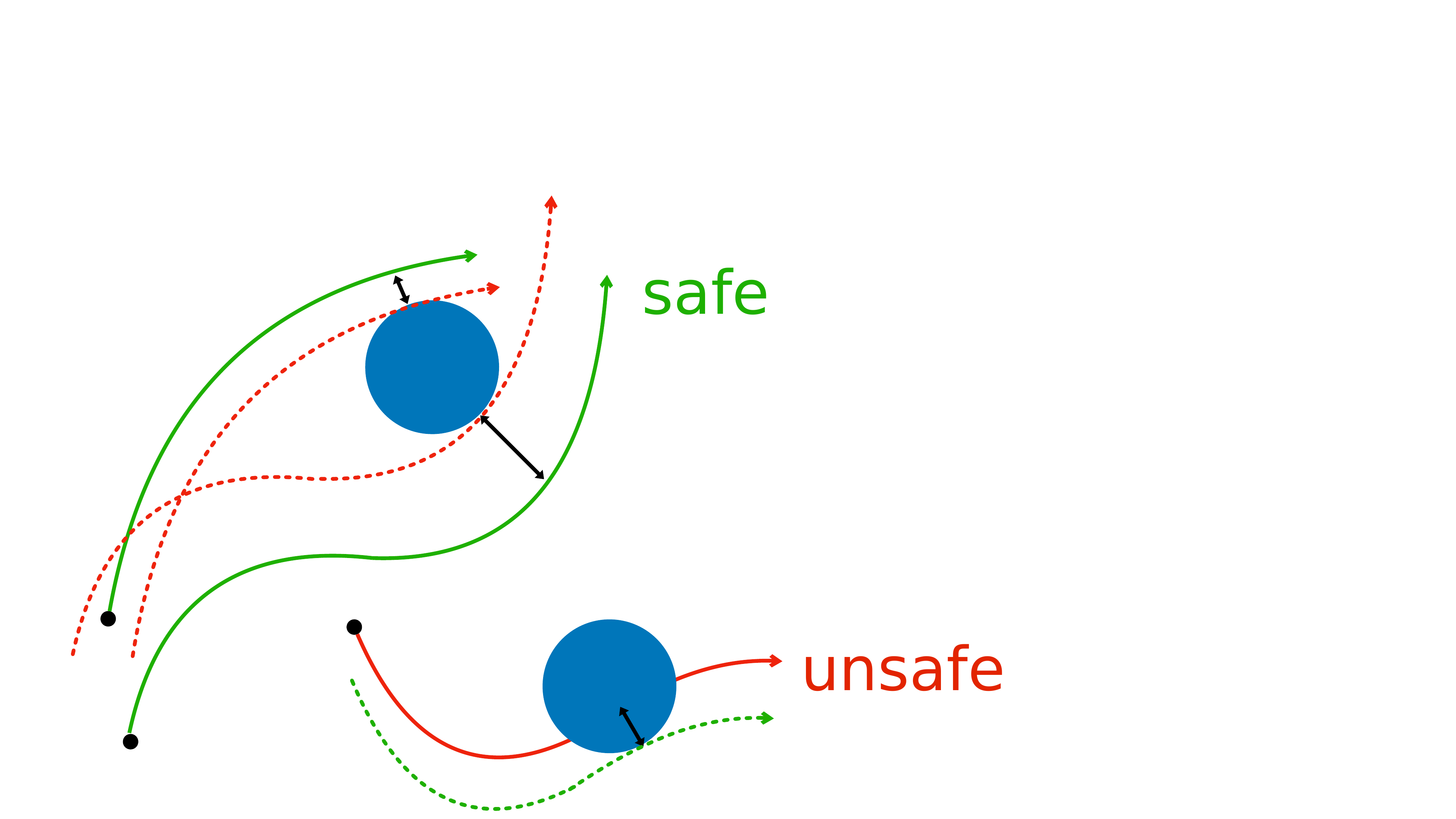}
    \hspace{-2.5cm}
    \includegraphics[scale=0.095]{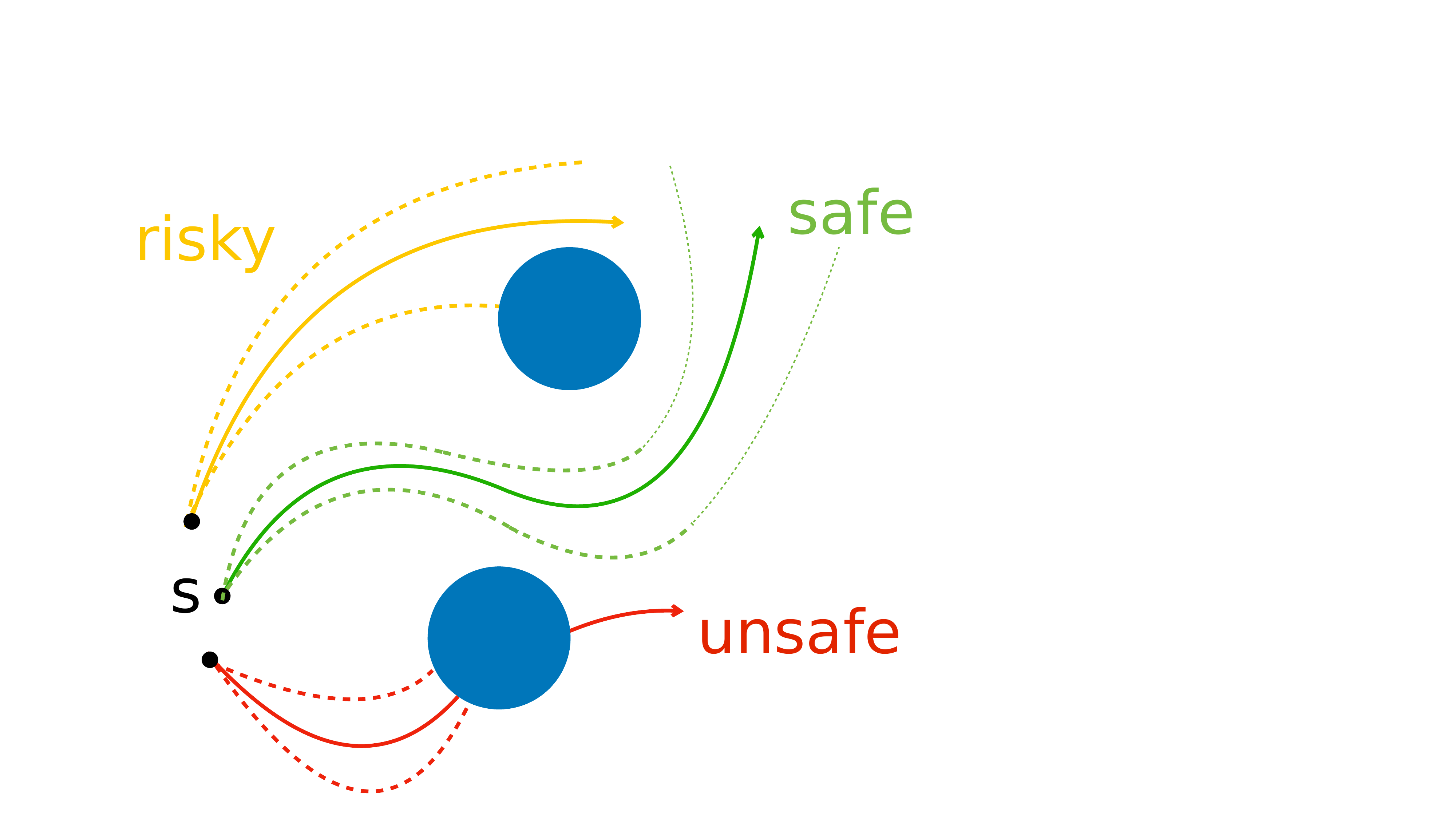}
    \hspace{-5cm}    
    \caption{Example of predictive monitoring of a safety property for a deterministic system (left) and a stochastic system (right). Blue circles denote the obstacles to avoid. (middle) shows the space robustness over the deterministic system.\vspace{-.5cm}}
    \label{fig:running}
\end{figure}

\paragraph{Running example.} Let's consider, as a running example, a point moving at a constant velocity on a two-dimensional plane (see Fig.~\ref{fig:running}). Given the system's current state $\state$, a controller regulates the yawn angle to avoid obstacles ($D_1$ and $D_2$ in our example). The avoid property can be easily expressed as an STL formula: $\varphi := G \left(\ (d(\state, o_1)> r_1) \land  (d(\state, o_2)> r_2)\ \right)$, where $o_i$, $r_i$ denote respectively the centre and the radius of obstacle $i\in\{1,2\}$. Fig.~\ref{fig:running} (left) shows the deterministic evolution for three randomly chosen initial states. Fig.~\ref{fig:running} (middle) shows, for the same deterministic scenario, an intuition of the concept of spatial STL robustness, i.e. how much we can perturb a trajectory with additive noise before changing its truth value. Fig.~\ref{fig:running} (right) shows the evolution of a stochastic dynamics for three randomly initial states. The dashed lines denote the upper and lower quantiles of the distribution over the trajectory space.

\vspace{-0.3cm}

\section{Predictive Monitoring}\label{sec:pm}

\vspace{-0.2cm}

Predictive monitoring of an HA is concerned with establishing whether given an initial state $\state$ and a desired property $\varphi$ -- e.g. always avoid a set of unsafe states $D$ -- the HA admits a trajectory starting from $\state$ that violates $\varphi$. 
We express such properties by means of time-bounded STL formulas, e.g. $\varphi_D := G_{[0, \FutureHorizon]}\ (s \not\in D)$. STL monitors automatically check whether an HA signal satisfies an STL property $\varphi$ over a bounded temporal horizon (Boolean semantics) and possibly how robust is the satisfaction (quantitative semantics).

\paragraph{SAT Oracles.} 
Given an HA $\HA$ with state space $\SM$, and an STL requirement $\varphi$ over a time bound $\FutureHorizon$, SAT oracles decide whether a state $\state\in\SM$ satisfies $\varphi$. 
This means that, when the system is \emph{deterministic}, the SAT oracle decides whether the unique trajectory $\Vec{\state}$ starting from $\state\in\SM$ satisfies $\varphi$, i.e. decide whether $(\Vec{\state}, 0) \models \varphi$. This information can be retrieved from STL monitors by checking whether $\chi^\varphi(\Vec{\state} ,0) = 1$ or equivalently whether $R_\varphi (\Vec{\state}, 0) > 0$. 
On the other hand, when the system is \emph{nondeterministic}, a state $\state$ satisfies $\varphi$ if all trajectories starting from $\state$  satisfy $\varphi$. Similarly, a quantitative SAT oracle returns the minimal STL robustness value of all trajectories starting from $\state$. 
For non-stochastic systems, 
the SAT oracle can thus be represented as a map $\sat: \SM\to\OutS$, where the output space $\OutS$ is either $\mathbb{B}$ in the Boolean scenario or $\mathbb{R}$ in the quantitative scenario. On the other hand, oracles for stochastic systems require a different treatment and we define them later in this section. 


\paragraph{SAT Tools.} Several tools have been developed for the automated verification of CPS properties and can thus be used as SAT oracles. The choice of the best tool depends on the problem at hand.
STL monitors such as Breach~\cite{donze2010breach} and RTAMT~\cite{nivckovic2020rtamt} allow to automatically check whether realizations of the system satisfy an STL property.
When the system is nondeterministic we need tools that perform reachability analysis or falsification. 
Due to the well-known undecidability of HA model checking problem~\cite{henzinger1995s,brihaye2011reachability}, none of existing tools are both sound and complete. Falsification tools like Breach~\cite{donze2010breach}, S-Taliro~\cite{annpureddy2011s}, C2E2~\cite{duggirala2015c2e2}, and HyLAA~\cite{bak2017hylaa} search for counter-example trajectories, i.e., for violations to the property of interest. A failure in finding a counter-example does not imply that the property is satisfied (i.e., the outcome is unknown). Conversely, HA reachability tools like PHAVer~\cite{frehse2005phaver}, SpaceEx~\cite{frehse2011spaceex}, Flow*, HyPro/HyDra~\cite{schupp2017h}, Ariadne~\cite{benvenuti2008reachability} and JuliaReach~\cite{bogomolov2019juliareach} rely on computing an over-approximation of the reachable set, meaning that the outcome is unknown when the computed reachable set intersects the target set. In order to be conservative, we treat unknown verdicts in a pessimistic way.

On the other hand, \emph{stochastic} systems require the use of probabilistic model checking techniques implemented in tools like PRISM~\cite{kwiatkowska2011prism} or STORM~\cite{hensel2021probabilistic}. Such tools provide precise numerical/symbolic techniques to determine the satisfaction probability of a formula, but only for a restricted class of systems and with significant scalability issues. Statistical model checking (SMC) techniques overcomes these limitations by solving the problem as one of hypothesis testing given a sample of system trajectories (at the cost of admitting some a priori statistical errors).

The above list of tools is far from being exhaustive, and we refer the interested reader to the ARCH-COMP competitions~\footnote{\url{https://cps-vo.org/group/ARCH/FriendlyCompetition}}, where state-of-the-art verification tools are compared on a set of well-known benchmarks.

\vspace{3mm}

In the following, we formulate the predictive monitoring problem for non-stochastic systems (Problem~\ref{prbl:pm}), for partially observable systems (Problem~\ref{prbl:pm_po}), and for stochastic systems (Problem~\ref{prbl:stoch_pm}). We conclude by characterizing the probabilistic guarantees sought for our learning-based monitors (Problem~\ref{prbl:stat_guar}).

We aim at deriving a predictive monitor for HA time-bounded satisfaction, i.e., a function that can predict whether or not the property $\varphi$ is satisfied by the future evolutions of the system (bounded by time $\FutureHorizon$) starting from the system's current state. In solving this problem, we assume a distribution $\StateDistrib$ of HA states and seek the monitor that predicts HA reachability with minimal error probability w.r.t.\ $\StateDistrib$. The choice of $\StateDistrib$ depends on the application at hand and can include a uniform distribution on a bounded state space or a distribution reflecting the density of visited states in some HA executions~\cite{phan2018neural}. 

\begin{myproblem}[Predictive monitoring for  HA]\label{prbl:pm}
Given an HA $\HA$ with state space $\SM$, a distribution $\StateDistrib$ over $\SM$, 
a time bound $\FutureHorizon$ and STL property $\varphi$, inducing the satisfaction function $\sat$, find a function $h^*: \SM \to \OutS$ that minimizes the probability 
$$\it Pr_{\state \sim \StateDistrib}\left( h^*(\state)\ne \sat(\state)\right).$$
A state $\state \in \SM$ is called \emph{positive} w.r.t.\ a predictor $h: \SM \to \OutS$ if $h(\state) > 0$. Otherwise, $\state$ is called \emph{negative}. 
\end{myproblem}

Any practical solution to the above PM problem must also assume a space of functions within which to restrict the search for the optimal predictive monitor $h^*$, for instance, one can consider functions described by deep neural networks (DNNs). 
Finding $h^*$, i.e., finding a function approximation with minimal error probability, is indeed a classical \textit{supervised learning} problem. In particular, in the Boolean scenario, $h^*$ is a classifier, i.e., a function mapping HA state inputs $\state$ into one of two classes: $1$ ($x$ is positive, property $\varphi$ is satisfied) and $0$ ($\state$ is negative, property $\varphi$ is violated). On the other hand, in the quantitative scenario, $h^*$ is a regressor aiming at reconstructing the robustness of satisfaction for a state $\state$. 
 
\begin{wrapfigure}[12]{r}{0.55\textwidth}
    \centering
    
    \vspace{-.7cm}
    
    \includegraphics[scale = 0.12]{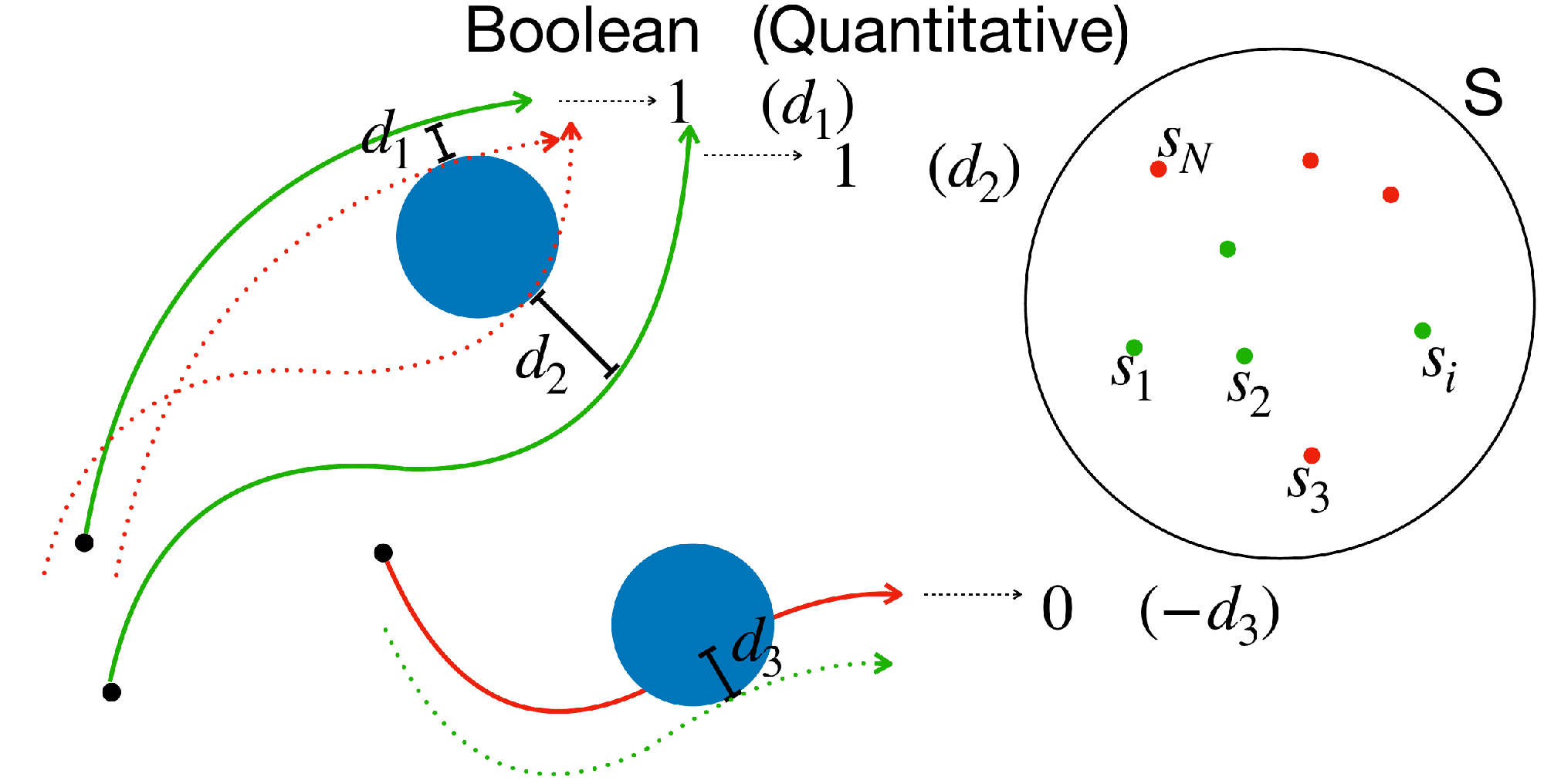}
    \caption{\footnotesize Generation of the dataset to learn a PM (Boolean or quantitative) for deterministic HS.\vspace{-1cm}}\label{fig:det_pm_dataset}

\end{wrapfigure}
\paragraph{Dataset generation.} In supervised learning, one minimizes a measure of the empirical prediction error w.r.t.\ a \textit{training set}. In our case, the training set $Z'$ is obtained from a finite sample $S'$ of $\StateDistrib$ by labelling the training inputs $\state \in S'$ using some SAT oracle, that is computing the true value for $\sat(\state)$. 
Hence, given a sample $S'$ of $\StateDistrib$, the training set is defined by 
$Z'=\{(\state,\ \sat (\state)) \mid \state \in S' \}$ (see Fig.~\ref{fig:det_pm_dataset}).

\subsubsection{Partial Observability}\label{sec:po}
Problem~\ref{prbl:pm} relies on the full observability (FO) assumption, i.e. the assumption of possessing full knowledge about the system's state. However, in most practical applications, state information is partial and noisy. 
Consider a discrete-time deterministic HS\footnote{In case of partial observability we restrict our analysis to deterministic systems.} modeled as a HA $\HA$. The discrete-time deterministic dynamics of the system can be expressed by $\cvar_{i+1} = Flow(\dvar_i)(\cvar_i)$, where $\state_i = (\dvar_i,\cvar_i) = (\dvar(t_i), \cvar(t_i))$ and $t_i = t_0+i\cdot\Delta t$. The measurement process can be modeled as 
\begin{equation}\label{eq:po}
y_i = \pi(\state_i)+w_i,    
\end{equation}
which produces partial and noisy observations $y_i \in Y$ by means of an  \emph{observation function} $\pi: \SM\to Y$ and  additive noise $w_i\sim \mathcal{W}$.
Under \emph{partial observability} (PO), we only have access to a sequence of past observations $\mathbf{y}_t = (y_{t-\PastHorizon},\dots,y_t)$ of the \textit{unknown} state sequence $\mathbf{\state}_t = (\state_{t-\PastHorizon},\dots,\state_t)$ (as per~\eqref{eq:po}). 
Let $\mathcal{Y}$ denote the distribution over $Y^{\PastHorizon}$, the space of sequences of observations $\mathbf{y}_t$ induced by the sequence of states $\mathbf{\state}_t\sim\StateDistrib^{\PastHorizon}$ and a sequence of i.i.d. noise vectors $\mathbf{w}_t = (w_{t-\PastHorizon},\dots,w_t) \sim \mathcal{W}^{\PastHorizon}$. 

\begin{myproblem}[PM for HS under noise and partial observability]\label{prbl:pm_po}
Given the HA and reachability specification of Problem~\ref{prbl:pm}, 
find a function $h_{po}^*: Y^{\PastHorizon} \rightarrow \OutS$ that minimizes 
\[{Pr}_{\state_t \sim \mathcal{S}, \mathbf{y}_t \sim \mathcal{Y}}
\Big(
h_{po}^*\big(\mathbf{y}_t\big)\ne \sat(\state_t) \big)
\Big).
\]
\end{myproblem}

In other words, $h_{po}^*$ should predict the satisfaction values given in input only a sequence of past observations, instead of the true HA state. 
In particular, we require a sequence of observations (as opposed to one observation only) for the sake of identifiability. Indeed, for general non-linear systems, a single observation does not contain enough information to infer the HS state\footnote{Feasibility of state reconstruction is affected by the time lag and the sequence length. Our focus is to derive the best predictions for fixed lag and sequence length, not to fine-tune these to improve identifiability.}. Problem~\ref{prbl:pm_po} considers only deterministic systems. Dealing with partial observability and noise in nondeterministic systems remains an open problem as state identifiability is a non-trivial issue.

There are two natural learning-based approaches to tackle Problem~\ref{prbl:pm_po} (Fig.~\ref{fig:nsc_diagram}):
\begin{enumerate}
    \item an \textbf{end-to-end} solution that learns a direct mapping from the sequence of past measurements $\mathbf{y}_t$ to the satisfaction value in $\OutS$.
    
    \item a \textbf{two-step} solution that combines steps (a) and (b) below:  
    \begin{itemize}
        \item[(a)] learns a \textit{state estimator} able to reconstruct the history of full states $\mathbf{s}_t = (s_{t-\PastHorizon},\dots,s_t)$ from the sequence of measurements $\mathbf{y}_t = (y_{t-\PastHorizon},\dots,y_t)$;
        \item[(b)] learns a \textit{state classifier/regressor} mapping the sequence of states $\mathbf{s}_t$ to the satisfaction value in $\OutS$;
    \end{itemize}
\end{enumerate}

\paragraph{Dataset generation.} Given that we consider deterministic dynamics, we can use simulation, rather than model checking, to label the states as safe (positive), if $\sat(\state)>0$, or unsafe (negative) otherwise. 
Because of the deterministic and Markovian (see Remark~\ref{remark:markov}) nature of the system, one could retrieve the future satisfaction of a property at time $t$ 
from the state of the system at time $t$ alone.  However, one can decide to exploit more information and make a prediction based on the previous $\PastHorizon$ states.
Formally, the generated dataset under FO can be expressed as 
$Z' = \{(\mathbf{s}^i_t, \sat(s^i_t))\}_{i=1}^N$, where $\mathbf{s}^i_t = (s^i_{t-\PastHorizon},s^i_{t-\PastHorizon+1},\dots, s^i_t)$.
Under PO, we use the (known) observation function $\pi:\SM\rightarrow Y$ to build a dataset 
 $Z''$ made of tuples $(\mathbf{y}_t, \mathbf{s}_t, l_t)$, where $\mathbf{y}_t$ is a sequence of noisy observations for $\mathbf{s}_t$, i.e., such that $\forall j\in\{t-\PastHorizon,\dots , t\}$ $y_{j} = \pi(s_j)+w_j$ and $w_j\sim\mathcal{W}$.
 
\tikzstyle{arrow} = [thick,->,>=stealth]
\tikzstyle{rect} = [rectangle, rounded corners, minimum width=3cm, minimum height=1cm,text centered, draw=black]

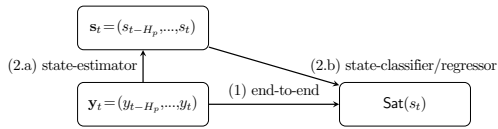
\begin{wrapfigure}[9]{r}{0.55\textwidth}
    \centering
    \vspace{-.5cm}
    \resizebox{0.55\textwidth}{!}{%
\begin{tikzpicture}

\node (meas) [rect] {$\mathbf{y}_t = ( y_{t-H_p},\dots , y_t) $};

\node (label) [rect, right of=meas, xshift=5cm] {$\sat(s_t)$};

\draw [arrow] (meas) -- node[anchor=south] {(1) end-to-end} (label);

\node (state) [rect, below of= meas, yshift=2.75cm] {$\mathbf{s}_t = (s_{t-H_p},\dots , s_t)$};

\draw [arrow] (state) -- node[anchor=west] {\qquad (2.b) state-classifier/regressor} (label);

\draw [arrow] (meas) -- node[anchor=east] {(2.a) state-estimator} (state);
\end{tikzpicture}
}%
    \caption{Diagram of the learning steps under noise and partial observability.\vspace{-1cm}}
    \label{fig:nsc_diagram}
\end{wrapfigure}

The distribution of $\mathbf{s}_t$ and $\mathbf{y}_t$ is determined by the distribution $\mathcal{S}$ of the initial state of the sequences, $s_{t-\PastHorizon}$. 
We consider two different distributions:  \emph{independent}, where the initial states $s_{t-\PastHorizon}$ are sampled independently, thus resulting in independent state/observation sequences; and 
\emph{sequential}, where states come from temporally correlated trajectories in a sliding-window fashion. The latter is more suitable for real-world runtime applications, where observations are received in a sequential manner. On the other hand, temporal dependency violates the exchangeability property, which affects the theoretical validity guarantees of CP, as we will soon discuss.

\subsubsection{Stochastic dynamics}
If the system evolves stochastically, we have a distribution over the trajectory space rather than a single trajectory that either satisfies of violates the property. Some realizations will satisfy the property, some others will not. Therefore, reasoning about satisfaction gets more complicated. 
Let $\mathbb{T} = \{0,1,\ldots\}$ denote a discrete set of time instants and let $\HA$ be a discrete-time stochastic HA over state space $\SM$ and $\mathbb{T}$. Given that the system is in state $\state \sim \StateDistrib$ at time $t \in \mathbb{T}$, the stochastic evolution (bounded by horizon $\FutureHorizon$) of the system starting at $\state$ can be described by the conditional distribution $p(\Vec{s} \mid \Vec{s}(t) = s),$
where $\Vec{s}=(\Vec{s}(t),\ldots, \Vec{s}(t+\FutureHorizon))\in S^H$ is the random trajectory of length $\FutureHorizon$ starting at time $t$. 
We thus introduce a satisfaction function $\stochsat$ that inherits the stochasticity of the system's dynamics. For an STL property $\varphi$, we define $\stochsat$ as a function mapping a state $\state\in\SM$ into a random variable $\stochsat(s)$ denoting the distribution of satisfaction values over $\OutS$. 
In other words, the satisfaction function transforms the distribution over trajectories into the distribution over satisfaction values.

The predictive monitoring problem under stochastic dynamics can be framed as estimating one or more functionals of $\stochsat(\state)$ (e.g., mean, variance, quantiles). 
A formal statement of the problem is given below. 

\begin{myproblem}[PM for Stochastic HS]\label{prbl:stoch_pm}
Given a discrete-time stochastic HA $\HA$ over a state space $\SM$,  temporal horizon $\FutureHorizon$, and an STL formula $\varphi$, we aim at approximating a functional $\mathsf{q}$ of the distributions induced by $\stochsat$. We thus aim at deriving a monitoring function $h_{\mathsf{q}}^*$ that maps any state $s \sim \StateDistrib$ into the functional $\mathsf{q}[\stochsat(s)]$ such that
\begin{equation}\label{eq:pi_validity}
    Pr_{\state\sim \StateDistrib}\Big(h_{\mathsf{q}}^*(\state) \ne \mathsf{q}\big[\stochsat(s)\big]\Big).
\end{equation}
\end{myproblem}

We will focus on the case where $\mathsf{q}$ is a quantile function, making Problem~\ref{prbl:stoch_pm} equivalent to a conditional quantile regression (QR) problem. This boils down to learning for a generic state $\state$ a quantile 
of the random variable $\stochsat(s)$.


\paragraph{Dataset generation.} 
We perform Monte-Carlo simulations of the process in order to obtain empirical approximations of $\stochsat$. 
In particular, we randomly sample $N$ states $s_1,\ldots,s_N\sim \StateDistrib$. Then, for each state $s_i$, we simulate $M$ trajectories of length $\FutureHorizon$, $\Vec{s}^1_i,\ldots , \Vec{s}^M_i$ where $\Vec{s}^j_i$ is a realization of $p(\Vec{s} \mid \Vec{s}(t) = s_i)$, and compute the satisfaction value $C_\varphi(\Vec{s}^j_i)$ of each of these trajectories ($C_\varphi\in\{\chi_\varphi,R_\varphi, Q_\varphi\}$). 
Note how $\{C_\varphi(\Vec{s}^j_i)\}_{j=1}^M$ is an empirical approximation of $\stochsat(s_i)$. 
The dataset is thus defined as 
\begin{equation}\label{eq:dataset}
    Z' = \Big\{\Big(s_i,\big(C_\varphi(\Vec{s}^1_i),\ldots,C_\varphi(\Vec{s}^M_i)\big)\Big), i=1,\ldots ,N \Big\}.
\end{equation}
\begin{wrapfigure}[14]{r}{0.6\textwidth}
    \centering
    \vspace{-0.7cm}
    \includegraphics[scale = 0.16]{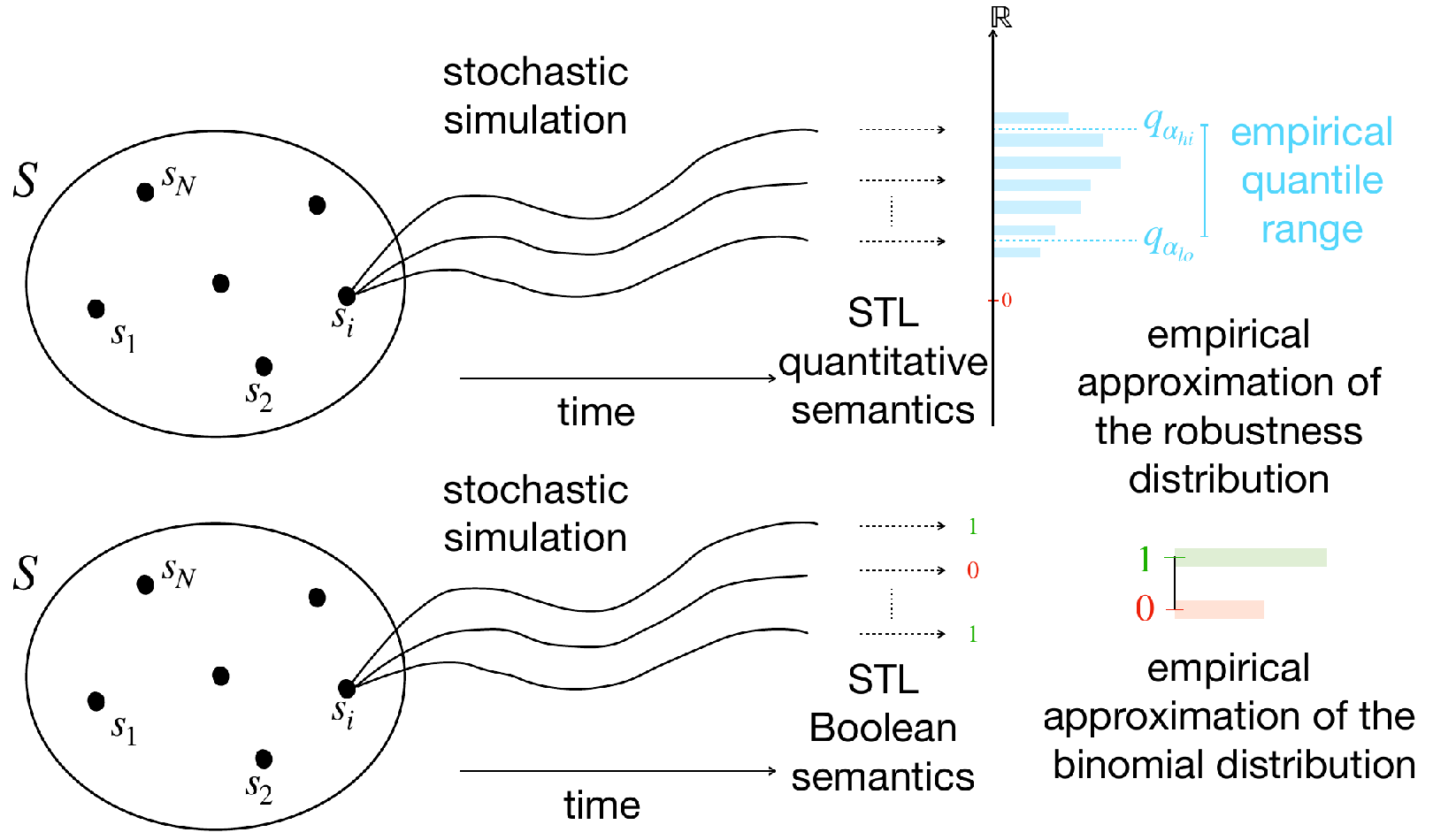}
    \caption{Generation of the dataset to learn a PM for stochastic HS.\vspace{-0.5cm}}
    \label{fig:stoch_pm_dataset}
\end{wrapfigure}
Fig.~\ref{fig:stoch_pm_dataset} shows an overview of the steps needed to generate the dataset.
The generation of the test set $Z'_{test}$ is very similar to that of $Z'$. The main difference is in that the number 
of trajectories that we simulate from each state $s$ is much larger than $M$. 
This allows us to obtain a highly accurate  
empirical approximation of the distribution induced by $\stochsat$, which we use as the ground-truth baseline in our experimental evaluation\footnote{In the limit of infinite sample size, the empirical approximation approaches the true distribution.}.
Moreover, since functionals of $\stochsat(\state)$ can not in general be computed exactly, for a choice of $\epsilon \in (0,1)$, we derive the empirical quantile $\hat{q}_{\epsilon}^{s_i}$ from samples $C_\varphi(\Vec{s}^1_i),\ldots,C_\varphi(\Vec{s}^M_i)$ and  use the generated training set $Z'$ to train the QR $h_{\mathsf{q}}$ that learns how to map states $s$ into $\hat{q}_{\epsilon}^s$.  

\vspace{3mm}

The predictors, either $h$, $h_{po}$ or $h_{\mathsf{q}}$, are approximate solutions and, as such, they can commit safety-critical prediction errors.
The general goal of Problems~\ref{prbl:pm}, \ref{prbl:pm_po} and~ \ref{prbl:stoch_pm} is to minimize the risk of making mistakes in predicting the satisfaction of a property. We are also interested in establishing probabilistic guarantees on the expected error rate of an \emph{unseen (test) state}, in the form of prediction regions guaranteed to include the true satisfaction value with arbitrary probability.
We now introduce some notation to capture all three previously stated scenarios. 
Let $f$ be the predictor (either $h$ of Problem~\ref{prbl:pm}, $h_{po}$ of Problem~\ref{prbl:pm_po} or $h_{\mathsf{q}}$ of Problem~\ref{prbl:stoch_pm}) and let $x\in X$ be the input of  predictor $f$ (either a state $\state$ or a sequence of past measurements $\mathbf{y}$). The distribution over the generic input space $X$ is denoted by $\mathcal{X}$.

\begin{myproblem}[Probabilistic guarantees]\label{prbl:stat_guar}
Given a system and property $\varphi$ as in Problems~\ref{prbl:pm},~\ref{prbl:pm_po} and ~\ref{prbl:stoch_pm}, 
find a function $\Gamma^{\varepsilon}: X\rightarrow 2^\OutS$, mapping every input $x$ into a prediction region for the corresponding satisfaction value, i.e., a region that satisfies, for any error probability level $\varepsilon \in (0,1)$, the \textit{validity} property below
\[
Pr_{x\sim \mathcal{X}}
\Big(
\barsat (x)\in \Gamma^{\varepsilon}(x)
\Big)\ge 1-\varepsilon,
\]
where $\barsat(\cdot)$ corresponds to $\sat(\cdot)$ in Problems~\ref{prbl:pm} and~\ref{prbl:pm_po}, and to $\mathsf{q}[\stochsat(\cdot)]$ in Problem~\ref{prbl:stoch_pm}.
\end{myproblem}
Among the maps that satisfy validity, we seek the most \emph{efficient} one, meaning the one with the smallest, i.e. less conservative, prediction regions.

\vspace{-0.3cm}

\section{Uncertainty Estimation and Statistical Guarantees}\label{sec:unc}

\vspace{-0.2cm}

The learning-based solutions of Problems~\ref{prbl:pm},~\ref{prbl:pm_po} and~\ref{prbl:stoch_pm} are approximate and, even when extremely high accuracies are reached, offer no guarantees over the reliability of the learned predictor, and thus are not applicable in safety-critical scenarios. 
In this section, we present techniques for uncertainty estimation, techniques that overcome the above limitation by providing point-wise information about the reliability of the predictions. In particular, we examine two uncertainty quantification techniques, based on conformal prediction (CP) and Bayesian inference, respectively. We focus more on CP as, unlike Bayesian inference, can provide the desired statistical guarantees stated in Problem~\ref{prbl:stat_guar}.

To simplify the presentation, we illustrate the techniques by considering a generic supervised learning model, as follows.
Let $X$ be the input space, $T$ be the target (output) space, and define $Z = X\times T$. Let $\mathcal{Z}$ be the data-generating distribution, i.e., the distribution of the points $(x,t)\in Z$.  
We assume that the target $t$ of a point $(x, t)\in Z$ is the result of the application of
a function $f^*:X\to T$, typically unknown or very expensive to evaluate. 
Using a finite set of observations, the goal of a supervised learning algorithm is to find a function $f: X\rightarrow T$ that accurately approximates $f^*$ over the entire input space. 
For a generic input $x\in X$, 
we denote with $t$ the true target value of $x$ and with $\hat{t}$ the prediction by $f$, i.e. $\hat{t} = f(x)$. 
Test inputs, whose unknown true target values we aim to predict, are denoted by $x_*$.

\subsection{Conformal Inference}\label{subsec:cp}


Conformal Prediction (CP) associates measures of reliability to any traditional supervised learning problem. It is a very general approach that can be applied across all existing deterministic classifiers and regressors~\cite{balasubramanian2014conformal,vovk2005algorithmic}. 
CP produces \textit{prediction regions with guaranteed validity}.

\begin{definition}[Prediction region]
For significance level $\varepsilon \in (0,1)$ and test input $x_*$, the $\varepsilon$-prediction region for $x_*$, $\Gamma_*^{\varepsilon}\subseteq T$, is a set of target values s.t.
\begin{equation}\label{eq:pred_r}
    \underset{(x_*,t_*)\sim \mathcal{Z}}{Pr}(t_* \in \Gamma_*^{\varepsilon}) = 1 - \varepsilon.
\end{equation}
\end{definition}

The idea of CP is to construct the prediction region by ``inverting'' a suitable hypothesis test: given a test point $x_*$ and a tentative target value $t'$, we \textit{exclude} $t'$ from the prediction region only if it is unlikely that $t'$ is the true value for $x_*$. The test statistic is given by a so-called \textit{nonconformity function (NCF)} $\ncf:Z \rightarrow \mathbb{R}$, which, given a predictor $f$ and a point $z=(x,t)$, measures the deviation between the true value $t$ and the corresponding prediction $f(x)$.  In this sense, $\ncf$ can be viewed as a generalized residual function. In other words, CP builds the prediction region $\Gamma_*^{\varepsilon}$ for a test point $x_*$ by excluding all targets $t'$ whose NCF values are unlikely to follow the NCF distribution of the true targets:
\begin{equation}\label{eq:cp_predr}
\Gamma_*^{\varepsilon} = \left\{t' \in T \mid  Pr_{(x,t)\sim \mathcal{Z}}\left(\ncf(x_*,t') \geq \ncf(x,t)\right) > \varepsilon\right\}.
\end{equation}
The probability term in Eq.~\ref{eq:cp_predr} is often called the p-value. 
From a practical viewpoint, the NCF distribution $Pr_{(x,t)\sim \mathcal{Z}}(\ncf(x,t))$ cannot be derived in an analytical form, and thus we use an empirical approximation derived using a sample $Z_c$ of $\mathcal{Z}$.  This  approach is called \emph{inductive} (or split) CP~\cite{papadopoulos2008inductive} and $Z_c$ is referred to as \textit{calibration set}.

\paragraph{Validity and Efficiency.} 
CP performance is measured via two quantities: 1) \emph{validity} (or \emph{coverage}), i.e. the empirical error rate observed on a test sample, which should be as close as possible to the significance level $\varepsilon$, and 2) \emph{efficiency}, i.e. the size of the prediction regions, which should be small. CP-based prediction regions are automatically valid, whereas the efficiency depends on the size of the calibration set (leading to high uncertainty when data is scarce), the quality of the underlying model and the chosen nonconformity function.

\begin{remark}[Assumptions and guarantees of inductive CP] Importantly, CP prediction regions have \textit{finite-sample validity}~\cite{balasubramanian2014conformal}, i.e., they satisfy~\eqref{eq:pred_r} for any sample of $\mathcal{Z}$ (of reasonable size), and not just asymptotically. On the other hand, CP's theoretical guarantees hold under the \textit{exchangeability} assumption (a ``relaxed'' version of iid) by which the joint probability of any sample of $\mathcal{Z}$ is invariant to permutations of the sampled points. Independent observations are exchangeable but sequential ones are not (due to the temporal dependency). In such scenarios, some adaptations to conformal inference (see~\cite{stankeviciute2021conformal,zaffran2022adaptive}) are needed to recover and preserve validity guarantees.
\end{remark}

\subsubsection{CP for classification} 

In classification, the target space is a discrete set of possible labels (or classes) $T=\{t^1,\ldots,t^c\}$.  We represent the classification model as a function $f_d: X\rightarrow [0,1]^c$ mapping inputs into a vector of class likelihoods, such that the predicted class is the one with the highest likelihood\footnote{Ties can be resolved by imposing an ordering over the classes.}. 

The inductive CP algorithm for classification is divided into an offline phase, executed only once, and an online phase, executed for every test point $x_*$. In the offline phase (steps 1--3 below), we train the classifier $f$ and construct the calibration distribution, i.e., the empirical approximation of the NCF distribution. In the online phase (steps 4--5), we derive the prediction region for $x_*$ using the computed classifier and distribution.
\begin{enumerate}
    \item Draw sample $Z'$ of $\mathcal{Z}$. Split $Z'$ into training set $Z_t$ and calibration set $Z_c$.
    \item Train classifier $f$ using $Z_t$. Use $f_d$ to define an NCF $\delta$.
    \item Construct the calibration distribution by computing, for each $z_i \in Z_c$, the NCF score $\alpha_i = \delta(z_i)$.
    \item For each label $t^j \in T$, compute $\alpha_*^j = \delta(x_*,t^j)$, i.e., the NCF score for $x_*$ and $t^j$, and the associated p-value $p_*^{j}$:
    \begin{equation}\label{eq:smoothed_p}
p_*^{j}= \frac{|\{z_i\in Z_c \mid \alpha_i > \alpha_*^{j}\}|}{|Z_c|+1}+\theta\frac{|\{z_i\in Z_c \mid \alpha_i = \alpha_*^{j}\}|+1}{|Z_c|+1},
\end{equation}
where $\theta\in\mathcal{U}[0,1]$ is a tie-breaking random variable. 
\item Return the prediction region 
\begin{equation}\label{eq:pred_reg}
   \Gamma_*^{\varepsilon} = \{t^j\in T \mid p_*^{j}>\varepsilon\}.
\end{equation}

\end{enumerate}

In defining the NCF $\delta$, we should aim to obtain high $\delta$ values for wrong predictions and low $\delta$ values for correct ones. Thus, a natural choice in classification is to define 
\begin{equation}\label{eq:ncf_class}
\delta(x,t^j) = 1 - f_d^j(x),    
\end{equation}
 where $f_d^j(x)$ is the likelihood predicted by $f_d$ for class $t_j$. Indeed, if $t^j$ is the true target for $x$ and $f$ correctly predicts $t^j$, then $f_d^j(x)$ is high (the highest among all classes) and $\delta(x,t^j)$ is low; the opposite holds if $f$ does not predict $t^j$.

\paragraph{Prediction uncertainty.} A CP-based prediction region provides a set of plausible predictions with statistical guarantees, and as such, also captures the uncertainty about the prediction. Indeed, if CP produces a region $\Gamma_*^{\varepsilon}$ with more than one class, then the prediction for $x_*$ is \textit{ambiguous} (i.e., multiple predictions are plausible), and thus, potentially erroneous. Similarly, if $\Gamma_*^{\varepsilon}$ is empty, then there are no plausible predictions at all, and thus, none can be trusted. The only reliable prediction is the one where $\Gamma_*^{\varepsilon}$ contains only one class. In this case, $\Gamma_*^{\varepsilon} = \{\hat{t}_*\}$, i.e., the region only contains the predicted class, as stated in the following proposition.

\begin{proposition}
For the NCF function~\eqref{eq:ncf_class}, if $\Gamma_*^{\varepsilon} = \{t^{j_1}\}$, then $t^{j_1} = f(x_*)$.
\end{proposition}

The size of the prediction region is determined by the chosen significance level $\varepsilon$ and by the p-values derived via CP. Specifically, from Equation~\eqref{eq:pred_reg} we can see that, for levels  $\varepsilon_1\ge\varepsilon_2$, the corresponding prediction regions are such that $\Gamma^{\varepsilon_1}\subseteq \Gamma^{\varepsilon_2}$.  
It follows that, given a test input $x_*$, if $\varepsilon$ is lower than all its p-values, i.e. if $\varepsilon < \min_{j=1,\ldots,c} \ p_*^{j}$, then the region $\Gamma_*^{\varepsilon}$ contains all the classes, and $\Gamma_*^{\varepsilon}$ shrinks as $\varepsilon$ increases. 
In particular, $\Gamma_*^{\varepsilon}$ is empty when $\varepsilon \geq \max_{j=1,\ldots,c} \ p_*^{j}$.

In the classification scenario, CP introduces two additional point-wise measures of uncertainty, called confidence and credibility, defined in terms of two p-values, independently of the significance level $\varepsilon$. The intuition is that these two p-values identify the range of $\varepsilon$ values for which the prediction is reliable, i.e., $|\Gamma_*^{\varepsilon}|=1$. 

\begin{definition}[Confidence and credibility]\label{def:conf_cred}
Given a predictor $F$, the \textit{confidence} of a point $x_*\in X$, denoted by $1-\gamma_*$, is defined as $
1-\gamma_* = \sup \{1-\varepsilon : |\Gamma_*^{\varepsilon}| = 1\}$,
 and the \textit{credibility} of $x_*$, denoted by $\credibility_*$, is defined as $
\credibility_* = \inf \{\varepsilon : |\Gamma_*^{\varepsilon}|= 0\}$.
The so-called \textit{confidence-credibility interval} $[\gamma_*, \credibility_*)$ contains all the values of $\varepsilon$ such that $|\Gamma_*^{\varepsilon}| = 1$. 

\end{definition}

The confidence $1-\gamma_*$ is the highest probability value for which the corresponding prediction region contains only $\hat{t}_*$, and thus it measures how likely (according to the calibration set) our prediction for $x_*$ is. 
In particular, $\gamma_*$ corresponds to the second largest p-value. The {credibility} $\credibility_*$ is the smallest level for which the prediction region is empty, i.e., no plausible prediction is found by CP. It corresponds to the highest p-value, i.e., the p-value of the predicted class.  
\begin{wrapfigure}[9]{r}{0.375\textwidth}
    \centering
\vspace{-.8cm}
    \includegraphics[scale=0.26]{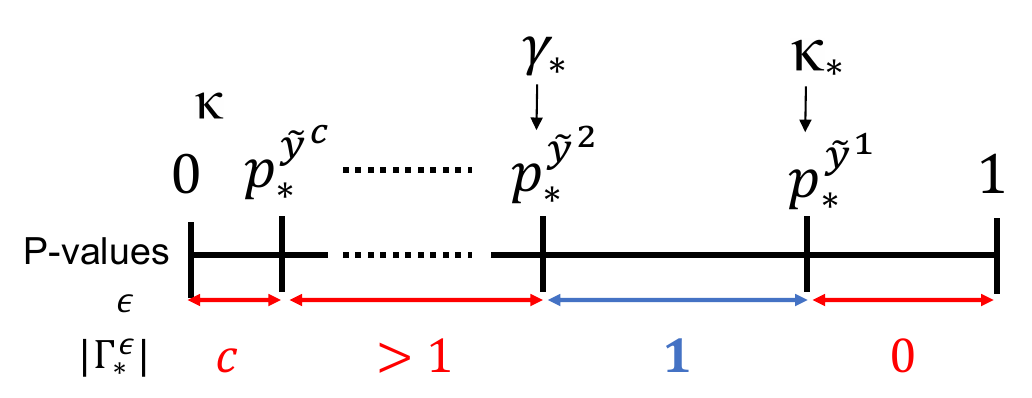}
\vspace{-8mm}
    \caption{{\footnotesize CP p-values and corresponding sizes of prediction interval. $\Tilde{y}^i$ is the class with the $i$-th largest p-value, so $p_*^{\Tilde{\ell}^1}=\credibility_*$ and  $p_*^{\Tilde{\ell}^2}=\gamma_*$. }}    \label{fig:pval}

\end{wrapfigure}
Fig.~\ref{fig:pval} illustrates CP p-values and corresponding prediction region sizes. In binary classification problems, each point $x_*$ has only two p-values: $\credibility_*$ (p-value of the predicted class) and $\gamma_*$ (p-value of the other class). 
It follows that the higher $1-\gamma_*$ and $\credibility_*$ are, the more reliable the prediction $\hat{t}_*$ is, because we have an expanded range $[\gamma_*, \credibility_*)$ of $\varepsilon$ values by which $|\Gamma_*^{\varepsilon}| = 1$.  
Indeed, in the degenerate case where $\credibility_*=1$ and $\gamma_*=0$, then $|\Gamma_*^{\varepsilon}| = 1$ for any value of $\varepsilon<1$. This is why, as we will explain in the next section,  we do not trust predictions with low values of $1-\gamma_*$ and $\credibility_*$. Hence, our CP-based uncertainty measure associates with each input its confidence and credibility values. 


\paragraph{Label-conditional approach.} The validity property, as stated above, guarantees an error rate over all possible labels, not on a per-label basis. 
The latter can be achieved with a CP variant, called
\emph{label-conditional CP}~\cite{gammerman2007hedging,shafer2008tutorial,toccaceli2019combination}. In this variant, the p-value associated with class $t^j$ on a test point $x_*$ is defined in a conditional manner as follows:
\begin{equation}\label{eq:mondrian_p}
p_*^{j}= \frac{|\{z_i\in Z_c : t_i=t^j, \alpha_i > \alpha_*^{j}\}|}{|\{z_i\in Z_{c}: t_i=t^j\}|+1}+ \theta\frac{|\{z_i\in Z_c : t_i=t^j, \alpha_i = \alpha_*^{j}\}|+1}{|\{z_i\in Z_{c}: t_i=t^j\}|+1}.
\end{equation}
In other words, we consider only the $\alpha_i$ corresponding to examples with the same
label $t^j$ as the hypothetical label that we are assigning at the test point.

Label-conditional validity is very important when CP is applied to an unbalanced dataset, whereby CP regions tend to have larger error rates with the minority class than with the majority one. 
The label-conditional approach ensures that, even for the
minority class, the expected error rate will tend to the chosen significance level $\varepsilon$.

\vspace{-0.25cm}

\subsubsection{CP for Regression}\label{sec:cp_regr}

In regression, we have a continuous target space $T\subseteq\mathbb{R}^n$. 
The CP algorithm for regression is similar to the classification one. In particular, the offline phase of steps 1--3, i.e., training of regression model $f$ and definition of NCF $\delta$, is the same (with obviously a different kind of $f$ and $\delta$). 

The online phase changes though, because $T$ is a continuous space and thus, it is not possible to enumerate the target values and compute for each a p-value. Instead, we proceed in an equivalent manner, that is, identify the critical value $\alpha_{(\varepsilon)}$ of the calibration distribution, i.e., the NCF score corresponding to a p-value of $\varepsilon$. The resulting $\varepsilon$-prediction region is given by $\Gamma_*^{\varepsilon} = f(x_*) \pm \alpha_{(\varepsilon)}$, where $\alpha_{(\varepsilon)}$ is the $(1-\varepsilon)$-quantile of the calibration distribution, i.e., the $\lfloor \varepsilon \cdot (|Z_c|+1)\rfloor$-th largest calibration score. 
A natural NCF in regression, and the one used in our experiments, is the norm of the difference between the real and the predicted target value, i.e., $\ncf(x, t) = ||t - f(x)||$. 

\paragraph{Normalized CP.}
The main limitation of CP for regression, presented above, is that the size of prediction intervals is identical ($2\alpha_{(\varepsilon)}$) for all test inputs, making CP non-informative to check how the uncertainty distributes over $X$.
Normalized Conformal Predictions (NCP)~\cite{Papadopoulos2011ReliablePI,Papadopoulos2014RegressionCP} tackle this limitation. In order to get individual, input-conditional bounds for each point $x_i$, we can define normalized nonconformity scores as follows
\begin{equation}\label{eq:norm_scores}
\tilde{\alpha}_c = \left\{\frac{\delta (x_i,t_i)}{u(x_i)}\ \Big| \ (x_i,t_i)\in Z_c\right\},
\end{equation}
where $u(x_i)$ estimates the difficulty of predicting $f(x_i)$. 
The rationale is that if two points have the same nonconformity scores using $\delta$, the one expected to be more accurate, should be stranger (more nonconforming) than the other one. Hence, we aim at error bounds that are tighter for inputs $x$ that are deemed easy to predict and vice-versa. Even for locally-weighted
residuals, as in~\eqref{eq:norm_scores}, the validity of the conformal methods carries over.
As before we compute $\tilde{\alpha}_{(\varepsilon)}$ as the $(1-\varepsilon)$-quantile of the scores $\tilde{\alpha}_c$ and the coverage guarantees over the error become:
\begin{equation}\label{eq:nicp}
    Pr_{(x,t)\sim\mathcal{Z}}\Big(
 \delta\big(x,t\big)\le\tilde{\alpha}_{(\varepsilon)}\cdot u(x)
    \Big) \ge 1-\varepsilon.
\end{equation}

\vspace{-0.25cm}

\paragraph{Conformalized Quantile Regression.}
The goal of conformalized quantile regression (CQR)~\cite{romano2019conformalized} is to adjust the QR prediction interval (i.e. the interval obtained by the prediction of two quantiles as in Problem~\ref{prbl:stoch_pm}) so that it is guaranteed to contain the $(1-\varepsilon)$ mass of probability.
As for CP, we divide the dataset $Z'$ in a training set $Z_t$ and a calibration set $Z_c$. We train the QR $f$ over $Z_t$ and on $Z_c$ we compute the nonconformity scores as
\begin{equation}\label{eq:CQR_cal_set}
\alpha_c := \max \{\hat{q}_{\varepsilon_{lo}}(x_i)-t_i, t_i-\hat{q}_{\varepsilon_{hi}}(x_i)\mid  (x_i,t_i)\in Z_c\}.
\end{equation}
In our notation, $\hat{q}_{\varepsilon_{lo}}(x)$ and $\hat{q}_{\varepsilon_{hi}}(x)$ denotes the two outputs of the pretrained predictor $f$ evaluated over $x$\footnote{If $f$ outputs more than two quantiles, $\hat{q}_{\varepsilon_{lo}}(x)$ and $\hat{q}_{\varepsilon_{hi}}(x)$ denote the predicted quantiles associated respectively with the lowest and highest associated significance level.}. 
The conformalized prediction interval is thus defined as
$$CPI(x_*) = [\hat{q}_{\varepsilon_{lo}}(x_*)-\alpha_{(\varepsilon)}, \hat{q}_{\varepsilon_{hi}}(x_*)+\alpha_{(\varepsilon)}],$$ where $\alpha_{(\varepsilon)}$ is the $\lfloor (1-\varepsilon)(1+1/|Z_c|)\rfloor$-th empirical quantile of $\alpha_c$. 
In the following, we will abbreviate with \textit{PI} a (non-calibrated) QR prediction interval and with \textit{CPI} a (calibrated) conformalized prediction interval.

Similarly to normalized CP, the above-defined CPI provides individual uncertainty estimates as the size of the interval changes according to the (input-conditional) quantile predictions.

\vspace{-0.25cm}

\begin{remark}
This nonconformity function, and thus $\alpha_{(\varepsilon)}$, can be negative and thus the conformalized prediction interval can be tighter than the original prediction interval.
This means that the CPI can be more efficient than the PI, where the efficiency is the average width of the prediction intervals over a test set.
The CPI has guaranteed coverage (the PI does not), i.e. $\mathbb{P}_{(x_*,t_*)\sim \mathcal{Z}}(t_*\in CPI(x_*))\ge 1-\varepsilon$.
\end{remark} 


\vspace{-0.25cm}

\paragraph{CP under Covariate Shift.}
CP guarantees hold under the assumption that training, calibration and test data come from the same data distribution $\mathcal{Z}$. However, there exist CP extensions~\cite{tibshirani2019conformal,cauchois2020robust} that provide statistical guarantees even in the presence of covariate shift at test time, meaning that the distribution $\mathcal{X}$ over inputs changes. The core concept is to reweight the nonconformity scores of the calibration set to account for the distribution shift. Such weights are defined using the density ratio between the shifted and original distribution, to quantify the probability of observing a particular calibration input relative to the shifted distribution.

\subsection{Bayesian Inference}\label{sec:bayes}

In general, a Bayesian inference problem aims at inferring an accurate probabilistic estimate of the unknown function from $X$ to $T$ (as before). In the following, let $f: X\to T$. The main ingredients of a Bayesian approach are the following:
\begin{enumerate}

    \item Choose a \emph{prior} distribution, $p(f)$, over a suitable function space, encapsulating the beliefs about function $f$ prior to any observations being taken.
    \item Determine the functional form of the observation process by defining a suitable \emph{likelihood} function $p(Z'| f)$ that effectively models how the  observations depend on the input.
    
    \item Leverage Bayes' theorem to define the \emph{posterior} distribution over functions given the observations
    $p(f| Z') = p(Z'| f)p(f)/p(Z')$.
    Computing $p(Z') = \int p(Z'| f) p(f) df$ is almost always computationally intractable as we have non-conjugate prior-likelihood distributions. Therefore, we need algorithms to accurately approximate such posterior distribution.
    
    \item Evaluate such posterior at points $x_*$, resulting in a predictive distribution $p(f_*|x_*,Z')$, whose statistics are used to obtain the desired estimate of the satisfaction probability together with the respective credible interval.

\end{enumerate}

\paragraph{Predictive uncertainty.} Once the empirical approximation of the predictive distribution $p(f_*|x_*,Z')$ is derived, one can extract statistics from it to characterize predictive uncertainty. We stress that the predictive distribution, and hence its statistics, effectively capture prediction uncertainty. For instance, the empirical mean and variance of the predictive distribution can be used as measures for Bayesian predictive uncertainty. 

\begin{remark}
    The Bayesian quantification of uncertainty, despite being based on statistically sound operations, offers no guarantees per se as it strongly depends on the chosen prior and typically relies on approximate inference. However, we can make predictions based on a functional of the predictive distribution and exploit the provided quantification of uncertainty as the normalizing function of an NCP approach, that in turn will provide us with point-specific statistical guarantees over the error coverage.
\end{remark}


In a Bayesian framework two main ingredients are essential to define the solution strategy to a Bayesian inference problem define above: 
$(i)$ the probabilistic model chosen to describe the distribution over functions $f$ and $(ii)$ the approximate inference strategy.
We refer to Appendix A of~\cite{bortolussi2022stochastic} for details on the possible approaches to Bayesian inference. In particular, we present Gaussian Processes and Bayesian Neural Nets, as alternatives for ingredient $(i)$, and Variational Inference (VI) and Hamilton Monte Carlo (HMC), as alternatives for ingredient $(ii)$. 

\vspace{-0.3cm}

\section{Learning-based PM with Statistical Guarantees}\label{sec:metods}


\subsection{Monitoring under Full Observability}

Given a fully observable Markovian system with a known SAT oracle, the system's current state $\state_t$ at time $t$ is sufficient information to predict the future satisfaction of a requirement $\varphi$. The input space of the $\sat$ function is thus  $\SM$. 

When the system evolves \emph{deterministically} each state $\state\in\SM$ is associated with a unique satisfaction value (as in Problem~\ref{prbl:pm}). If we are interested in the Boolean satisfaction the output space $\OutS$ of the $\sat$ function is $\{0,1\}$ and the learning problem is a classical binary classification problem, i.e. inferring a function $h:\SM\to\{0,1\}$ that classify a state as positive (if it satisfies the requirement) or negative (if it violates the requirement). Analogously, if we want to better quantify how robust is the satisfaction we can leverage the quantitative STL semantics, either spatial or temporal. In this scenario, the output space $\OutS$ of the $\sat$ function is $\mbb{R}$ and the learning problem becomes a regression task, i.e. inferring a function $h:\SM\to\mbb{R}$ that estimates the level of satisfaction of $\varphi$ for each state in $\SM$. 

The function $h$, introduced in its general form in Problem~\ref{prbl:pm}, can be inferred either using a deterministic neural network or one of the proposed Bayesian approaches. CP can be used on top of both approaches to meet the validity guarantees of Problem~\ref{prbl:stat_guar}. In the CP-based version, we apply either CP for classification or CP for regression to obtain prediction regions with guaranteed coverage over the entire state space (see~\cite{bortolussi2019neural} for details). On the other, we can design either a Bayesian classifier (with Bernoulli likelihood for the Boolean semantics) or a Bayesian regressor (Gaussian likelihood for quantitative semantics). See~\cite{bortolussi2021neural} for details. In order to meet the desired statistical guarantees we could use CP. Since Bayesian predictions are probabilistic, whereas CP is defined for deterministic predictors, we apply CP to the expectation over the predictive distribution. Nonetheless, the variance of the latter can be exploited as normalizing constant in a NCP framework so to obtain state-specific prediction intervals while preserving the statistical guarantees. Similar reasoning is applied to \emph{nondeterministic systems}.

\begin{figure}[!b]
    \centering

    \vspace{-0.75cm}
    
    \includegraphics[scale=0.075]{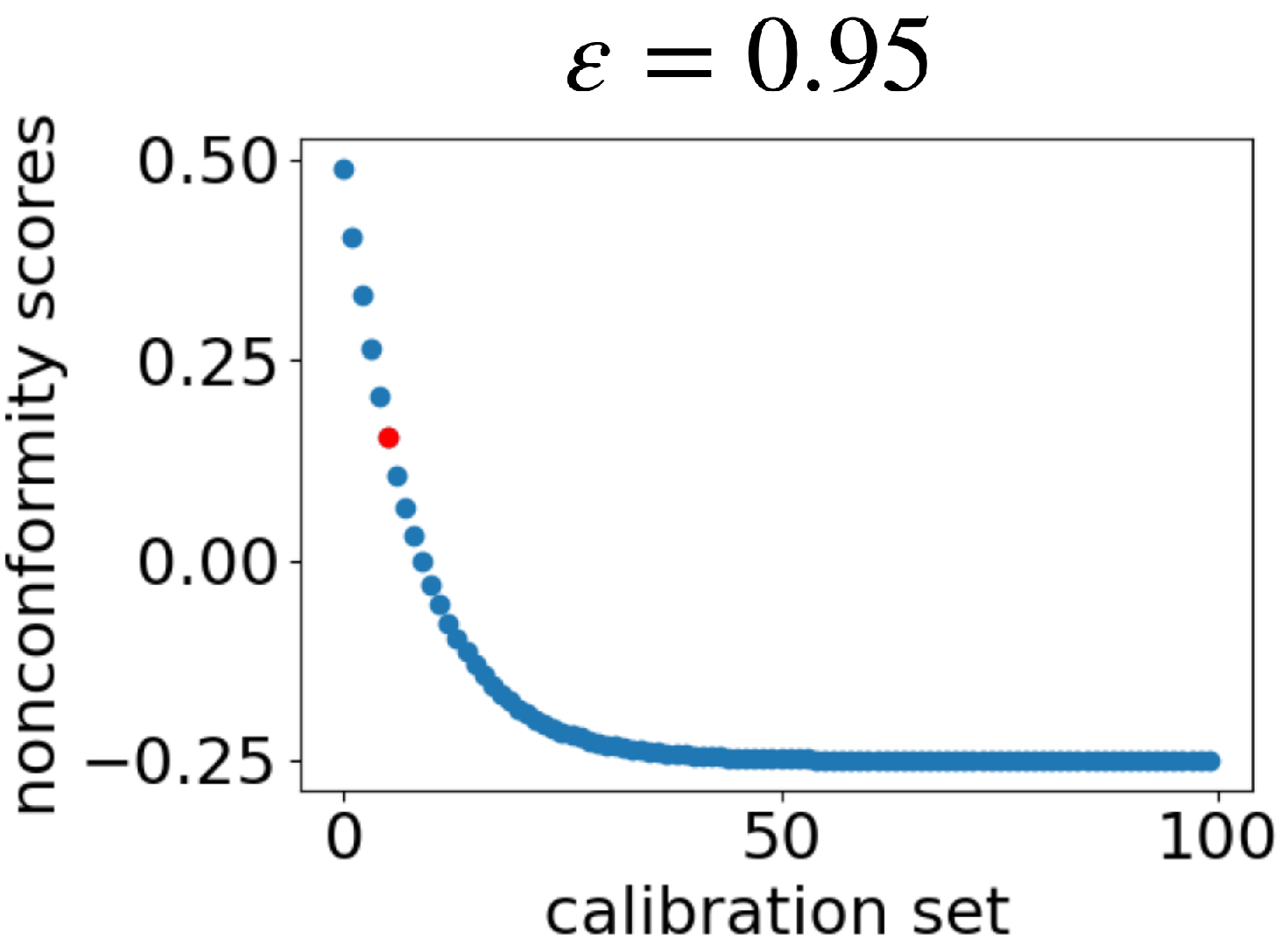}
    \includegraphics[scale=0.1]{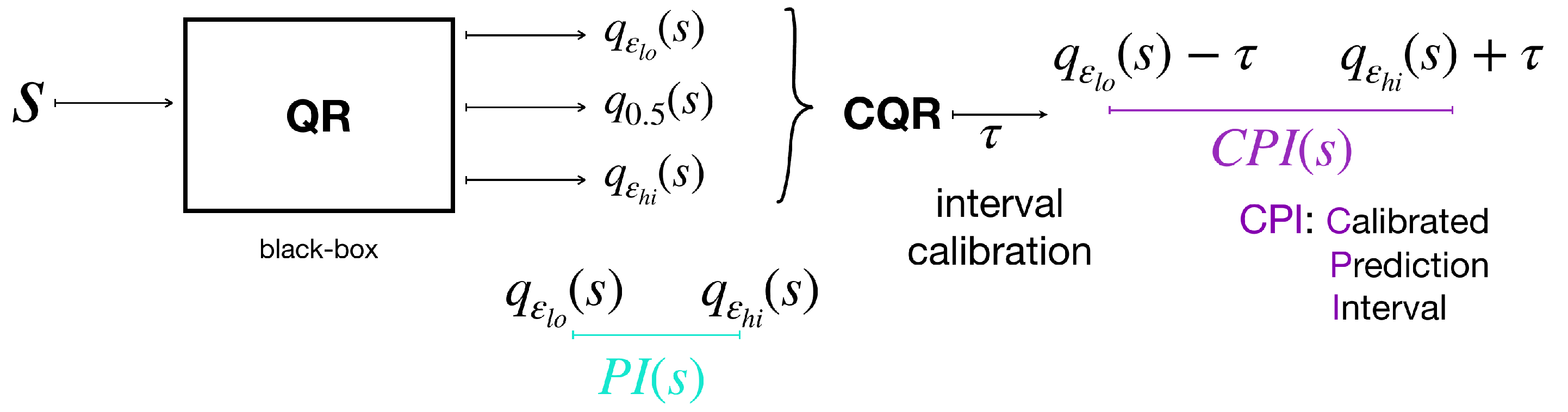}
    \vspace{-0.4cm}
    \caption{Overview of conformalized quantile regression.\vspace{-1cm}}
    \label{fig:cqr}
\end{figure}

Otherwise, when the system evolves \emph{stochastically}, each state $\state\in\SM$ is associated with a distribution over the satisfaction values $\stochsat(\state)$ (as discussed in  Problem~\ref{prbl:stoch_pm} in Section~\ref{sec:pm}). We are not able to extract an analytic expression for this distribution but we can empirically approximate it via sampling. If we consider the Boolean semantics, $\stochsat(\state)$ is a Binomial distribution centred around the satisfaction probability in the interval $[0,1]$. In such a scenario, we could either train a deterministic neural regressor that infers the satisfaction probability in $[0,1]$ or design a Bayesian framework with Binomial likelihood (see~\cite{bortolussi2022stochastic} for details). In order to meet the desired statistical guarantees we could use CP for regression. Once again, in the Bayesian scenario, CP is applied to the expectation over the predictive distribution. However, the variance of the latter can be used as normalizing constant in a NCP framework so to obtain state-specific prediction intervals.
On the other hand, the quantitative STL semantics, either spatial or temporal, results in a distribution over $\mbb{R}$. We can train neural quantile regression (QR) that, given a desired confidence level $\varepsilon$, infers some quantiles of this distribution (e.g. $q_{\tfrac{\varepsilon}{2}}$, $q_{0.5}$ and $q_{\tfrac{\varepsilon}{2}}$). A typical loss for the regression of a quantile $q_\alpha$ is the pinball loss $
\mathcal{L}_\alpha(t,\hat{q}_\alpha) = \alpha\cdot \max(t-\hat{q}_\alpha,0)+(1-\alpha)\cdot\max(\hat{q}_\alpha-t,0),
$
where $\hat{q}_\alpha$ is the predicted quantile and $t\in\mbb{R}$ denotes an observed output. Once the QR is trained we can resort to CQR (see Fig.~\ref{fig:cqr}) to meet the probabilistic guarantees, in that the
conformal intervals cover with probability at least $1-\varepsilon$ the STL robustness values
relative to the stochastic evolution of the system. The rationale is to evaluate the nonconformity scores of the interval $\left[\hat{q}_{\tfrac{\varepsilon}{2}}(\state), \hat{q}_{1-\tfrac{\varepsilon}{2}}(\state)\right]$ over the calibration set and extract $\tau$, the $\lfloor (1-\varepsilon)(1+1/|Z_c|)\rfloor$-th empirical quantile of $\alpha_c$, to recalibrate the prediction interval (see~\cite{cairoli2023conformal} for details).

\vspace{-0.3cm}

\subsection{Monitoring under Partial Observability }
For ease of discussion, in the PO scenario (outlined in Problem~\ref{prbl:pm_po}), we discuss only the CP-based setting and not the Bayesian one (see~\cite{cairoli2021neural} for details). 
The \emph{end-to-end} approach is very similar to the FO deterministic one. The main difference is that instead of the state at time $t$ we map the history of past measurements $\mathbf{y}_t$ to the satisfaction value in $\OutS$, i.e. we infer a function $h_{po}:Y^{\PastHorizon}\to\OutS$. As before, the output can be either Boolean $\OutS = \{0,1\}$ (binary classification task), or quantitative $\OutS = \mathbb{R}$ (regression task). The sequence of past observations is mapped to a unique satisfaction value in $\OutS$ and CP can be used to enrich the predictions with guaranteed validity.
On the other hand, if we consider a \emph{two-step} approach we first estimate the sequence of states $\mathbf{s}_t$ (regression task) and then we estimate the satisfaction value associated with each sequence which is either a classification or a regression task (as in the end-to-end approach). The two steps can be fine-tuned together and conformal inference can be applied to both steps to obtain statistical guarantees.

\vspace{-0.3cm}

\subsection{Uncertainty-aware Error Detection and Active Learning}

It is well known that neural networks are universal approximators. However, such methods cannot completely avoid prediction errors (no supervised learning method can).  
Therefore, we have to deal with predictive monitors $f$ that are prone to prediction errors: when, for a state $s \in \SM$, $f(x)\ne\sat(x)$. These errors are respectively denoted by predicates $\mathit{pe}(\state)$. 

\begin{myproblem}[Uncertainty-based error detection]\label{prbl:ed}
Given a reachability predictor $f$, a distribution $\mathcal{X}$ over HA states $X$, a predictive uncertainty measure $u_f: X\to U$ over some  uncertainty domain $U$, and a kind of error $\mathit{pe}$
find an optimal error detection rule ${G}_{f,pe}^*: U \to  \{0,1\}$, i.e., a function that minimizes the probability $$\it Pr_{x\sim \mathcal{X}} \left(pe(x) \neq {G}^*_{f,pe}(u_f(x))\right).$$ 
\end{myproblem}
In the above problem, we consider all kinds of prediction errors, but the definition and approach could be easily adapted to focus on the detection of only e.g., false positives (the most problematic errors from a safety-critical viewpoint). 

In the CP-based setting, a meaningful measure of predictive uncertainty is given by  confidence and credibility. In the Bayesian framework, we can consider the mean and the variance of the predictive distribution.

As for Problem~\ref{prbl:pm},~\ref{prbl:pm_po} and~\ref{prbl:stoch_pm}, we can obtain a sub-optimal solution ${G}_{f,pe}$ to Problem~\ref{prbl:ed} by expressing the latter as a supervised learning problem, where the inputs are, once again, sampled according to $\mathcal{X}$ and labelled using a SAT oracle. We call \emph{validation set} the set of labelled observations used to learn ${G}_{f,pe}$.
These observation need to be independent from the above introduced training set $Z'$, i.e., those used to learn the reachability predictor $f$. 
The final rejection rule $\mathsf{Rej}_{f,pe}$ for detecting HA states where the satisfaction prediction (given by $f$) should not be trusted, and thus rejected, is readily obtained by the composition of the uncertainty measure and the error detection rule 
$\mathsf{Rej}_{f,pe} = {G}_{f,e}\circ u_f: X\to \{0, 1\},$
where $\mathsf{Rej}_{f,pe}(x)=1$ if the prediction on $x$ is rejected and $\mathsf{Rej}_{f,pe}(x)=0$ otherwise.

This error-detection criterion can be also used as a query strategy in an uncertainty-aware \emph{active learning} setting. Active learning should reduce the overall number of erroneous predictions because it improves
the predictor on the inputs where it is most uncertain.

\vspace{-0.3cm}

\section{Related Work}\label{sec:related}

\vspace{-0.2cm}

A number of methods have been proposed for online reachability analysis that rely on separating the reachability computation into distinct offline and online phases. However, these methods are limited to restricted classes of models~\cite{chen2017model,yoon2019predictive}, or require handcrafted optimization of the HA's derivatives~\cite{bak2014real}, or are efficient only for low-dimensional systems and simple dynamics~\cite{sauter2009lightweight}. 
In contrast, the approaches presented in this paper are based on learning DNN-based predictors, are fully automated and have negligible computational cost at runtime. In~\cite{djeridane2006neural,royo2018classification}, similar techniques are introduced for neural approximation of Hamilton-Jacobi (HJ) reachability. 
However, our methods for prediction rejection and active learning are independent of the class of systems and the machine-learning approximation of reachability, and thus can also be applied to neural approximations of HJ reachability. 
In~\cite{yel2020assured}, Yel and others present a runtime monitoring framework that has similarities with our approach, in that they also learn neural network-based reachability monitors (for UAV planning applications), but instead of using,  like we do,  uncertainty measures to pin down potentially erroneous predictions, they apply NN verification techniques~\cite{ivanov2019verisig} to identify input regions that might produce false negatives. Thus, their approach is complementary to our uncertainty-based error detection, but, due to the limitations of the underlying verification algorithms, they can only support deterministic neural networks with sigmoid activations. On the contrary, our techniques support any kind of ML-based monitors, including probabilistic ones.
The work of~\cite{babaee2018predictive,babaee2019accelerated} addresses the predictive monitoring problem for stochastic black-box systems, where a Markov model is inferred offline from observed traces and used to construct a predictive runtime monitor for probabilistic reachability checking. In contrast to our method, this method focuses on discrete-space models, which allows the predictor to be represented as a look-up table, as opposed to a neural network.
In~\cite{qin2019predictive}, a method is presented for predictive monitoring of STL specifications with probabilistic guarantees. These guarantees derive from computing prediction intervals of ARMA/ARIMA models learned from observed traces. Similarly, we use CP which also can derive prediction intervals with probabilistic guarantees, with the difference that CP supports any class of prediction models (including auto-regressive ones). In~\cite{dokhanchi2014line}, model predictions are used to forecast future robustness values of MTL specifications for runtime monitoring. However, no guarantee, statistical or otherwise, is provided for the predicted robustness. Deshmukh and others~\cite{deshmukh2017robust} have proposed an interval semantics for STL over partial traces, where such intervals are guaranteed to include the true STL robustness value for any bounded continuation of the trace. This approach can be used in the context of predictive monitoring but tends to produce over-conservative intervals.
Another related approach is smoothed model checking~\cite{bortolussi2016smoothed}, where Gaussian processes~\cite{rasmussen2006gaussian} are used to approximate the satisfaction function of stochastic models, i.e., mapping model parameters into the satisfaction probability of a specification. Smoothed model checking leverages Bayesian statistics to quantify prediction uncertainty, but faces scalability issues as the dimension of the system increases. These scalability issues are alleviated in~\cite{bortolussi2022stochastic} using stochastic variational inference.
In contrast, computing our conformal measure of prediction reliability is very efficient, because it is nearly equivalent to executing the underlying predictor.

This tutorial builds on the methods presented in~\cite{phan2018neural,bortolussi2019neural,bortolussi2021neural,cairoli2021neural,cairoli2022neural,cairoli2023conformal,bortolussi2022stochastic}. In NPM~\cite{bortolussi2019neural,bortolussi2021neural}, neural networks are used to infer the Boolean satisfaction of a reachability property and conformal prediction (CP) are used to provide statistical guarantees. 
NPM has been extended to support some source of stochasticity in the system: in~\cite{cairoli2021neural} they allow partial observability and noisy observations, in~\cite{cairoli2022neural} the system dynamics are stochastic but the monitor only evaluates the Boolean satisfaction of some quantile trajectories, providing a limited understanding of the safety level of the current state. Finally in~\cite{cairoli2023conformal} a conformal quantitative predictive monitor to reliably check the satisfaction of STL requirements over evolutions of a stochastic system at
runtime is presented.
Predictive monitoring under partial observability is also analysed in~\cite{chou2020predictive},  where the authors combine Bayesian state estimation with pre-computed reach sets to reduce the runtime overhead. While their reachability bounds are certified, no correctness guarantees can be established for the estimation step.

Various learning-based PM approaches for temporal logic properties~\cite{qin2020clairvoyant,ma2021predictive,yoon2021predictive,yu2022model,rodionova2021time,muthali2023multi} have been recently proposed. 
In particular, Ma et al.~\cite{ma2021predictive} use uncertainty quantification with Bayesian RNNs to provide confidence guarantees. However, these models are, by nature, not well-calibrated (i.e., the model uncertainty does not reflect the observed one~\cite{kuleshov2018accurate}), making the resulting guarantees not theoretically valid. 
In~\cite{badings2022sampling} the parameter space of a parametric CTMC is explicitly explored, while~\cite{vcevska2017precise} assumes a probability distribution over the parameters and proposes a sampling-based approach. In~\cite{lindemann2023conformal} conformal predictions are used over the expected value of the stochastic process rather than its distribution.

We contribute to the state of the art by presenting a wide variety of learning-based predictive monitors that offer good scalability, provide statistical guarantees, and support partial observability, stochasticity and rich STL-based requirements.

\vspace{-0.3cm}

\section{Conclusions}\label{sec:conclusion}

\vspace{-0.2cm}

We have presented an overview of various learning-based approaches to reliably monitor the evolution of a CPS at runtime. The proposed methods complement predictions over the satisfaction of an STL specification with principled estimates of the prediction uncertainty. These estimates can be used to derive optimal rejection criteria that identify potentially erroneous predictions without knowing the true satisfaction values. The latter can be exploited as an active learning strategy increasing the accuracy of the satisfaction predictor. The strength is given by high-reliability and high computational efficiency of our predicitons. The efficiency is not directly affected by the complexity of the system under analysis but only by the complexity of the learned predictor. 
Our approach overcomes the computational footprint of model checking (infeasible at runtime) while improving on traditional runtime verification by being able to detect future violations in a preemptive way.   
We have devised two alternative solution methods: a frequentist and a Bayesian approach. Conformal predictions are used on top of both methods to obtain statistical guarantees.

In future work, 
we will investigate dynamics-aware approaches to inference. The aim is to improve the performances by limiting inference only to an estimate of the system manifold, i.e. the region of the state space that is likely to be visited by the evolving stochastic process.

\vspace{5mm}

\noindent \textbf{Acknowledgments.} 
This work has been partially supported by the PRIN project ``SEDUCE'' n. 2017TWRCNB, by the ``REXASI-PRO'' H-EU project, call HORIZON-CL4-2021-HUMAN-01-01, Grant agreement ID: 101070028 and by the PNRR project iNEST (Interconnected North-Est Innovation Ecosystem) funded by the European Union Next-GenerationEU (Piano Nazionale di Ripresa e Resilienza (PNRR) – Missione 4 Componente 2, Investimento 1.5 – D.D. 1058 23/06/2022, ECS\_00000043).

\bibliographystyle{splncs04}
\bibliography{biblio}


\end{document}